\documentclass[10pt,twocolumn,letterpaper]{article}
\usepackage{cvpr}              

\usepackage[accsupp]{axessibility} 
\usepackage{times}
\usepackage{epsfig}
\usepackage{graphicx}
\usepackage{amsmath}
\usepackage{amssymb}
\usepackage{adjustbox}
\usepackage{booktabs}

\usepackage[pagebackref=true,breaklinks=true,colorlinks,bookmarks=false]{hyperref}

\def\confYear{NTIRE 2022}

\begin{document}
    
\title{NTIRE 2022 Challenge on High Dynamic Range Imaging:\\Methods and Results}

\author{Eduardo P\'erez-Pellitero$^\dagger$ \and Sibi Catley-Chandar$^\dagger$ \and Richard Shaw$^\dagger$ \and Ale\v{s} Leonardis$^\dagger$ \and Radu Timofte$^\dagger$ \and
Zexin Zhang \and  Cen Liu \and  Yunbo Peng\and  Yue Lin \and 
Gaocheng Yu \and Jin Zhang\and Zhe Ma\and Hongbin Wang \and 
Xiangyu Chen \and Xintao Wang \and Haiwei Wu \and Lin Liu \and Chao Dong \and Jiantao Zhou \and Qingsen Yan \and Song Zhang \and Weiye Chen \and Yuhang Liu \and Zhen Zhang \and Yanning Zhang \and Javen Qinfeng Shi \and Dong Gong \and 
Dan Zhu \and Mengdi Sun \and Guannan Chen \and
Yang Hu \and Haowei Li \and Baozhu Zou \and Zhen Liu \and Wenjie Lin \and Ting Jiang \and Chengzhi Jiang \and  Xinpeng Li \and Mingyan Han \and Haoqiang Fan \and Jian Sun \and Shuaicheng Liu \and
Juan Mar\'in-Vega \and Michael Sloth \and Peter Schneider-Kamp \and Richard R\"ottger \and  
Chunyang Li \and Long Bao \and 
Gang He \and Ziyao Xu \and Li Xu \and Gen Zhan \and Ming Sun \and Xing Wen \and Junlin Li \and  
Jinjing Li \and Chenghua Li \and Ruipeng Gang \and Fangya Li \and Chenming Liu \and Shuang Feng \and  
Fei Lei \and Rui Liu \and Junxiang Ruan \and
Tianhong Dai \and Wei Li \and Zhan Lu \and Hengyan Liu \and Peian Huang \and Guangyu Ren \and 
Yonglin Luo \and Chang Liu \and Qiang Tu \and 
Fangya Li \and Ruipeng Gang \and Chenghua Li \and Jinjing Li \and Sai Ma \and Chenming Liu \and Yizhen Cao \and
Steven Tel \and Barthelemy Heyrman \and Dominique Ginhac \and
Chul Lee \and Gahyeon Kim \and Seonghyun Park \and An Gia Vien \and Truong Thanh Nhat Mai \and 
Howoon Yoon \and Tu Vo \and Alexander Holston \and Sheir Zaheer \and Chan Y.\, Park 
}
\maketitle

{\let\thefootnote\relax\footnotetext{%
\hspace{-5mm}$\dagger$ Eduardo P\'erez-Pellitero (\texttt{e.perez.pellitero@huawei.com}), Sibi Catley-Chandar, Richard Shaw, Ale\v{s} Leonardis (Huawei Noah's Ark Laboratory) and Radu Timofte (ETH Z\"urich, University of W\"urzburg) are the NTIRE 2022 Challenge organizers, while the other authors participated in the challenge. Appendix~\ref{ap:teams-and-affiliations} contains the authors’ teams and affiliations.\\
\url{https://data.vision.ee.ethz.ch/cvl/ntire22/}}}
\begin{abstract}
This paper reviews the challenge on constrained high dynamic range (HDR) imaging that was part of the New Trends in Image Restoration and Enhancement (NTIRE) workshop, held in conjunction with CVPR 2022. This manuscript focuses on the competition set-up, datasets, the proposed methods and their results. The challenge aims at estimating an HDR image from multiple respective low dynamic range (LDR) observations, which might suffer from under- or over-exposed regions and different sources of noise. The challenge is composed of two tracks with an emphasis on fidelity and complexity constraints: In Track 1, participants are asked to optimize objective fidelity scores while imposing a low-complexity constraint (\ie solutions can not exceed a given number of operations). In Track 2, participants are asked to minimize the complexity of their solutions while imposing a constraint on fidelity scores (\ie solutions are required to obtain a higher fidelity score than the prescribed baseline). Both tracks use the same data and metrics: Fidelity is measured by means of PSNR with respect to a ground-truth HDR image (computed both directly and with a canonical tonemapping operation), while complexity metrics include the number of Multiply-Accumulate (MAC) operations and runtime (in seconds).
\end{abstract} 

\section{Introduction}
Advances in computational photography have enabled single-sensor cameras to acquire images with an extended dynamic range without the need for expensive, bulky and arguably more inconvenient multi-camera rigs, \eg~\cite{Tocci11, McGuire07, Froehlich14}. The principle behind such single-sensor camera designs relies upon multiple conventional frames captured with different exposure values (EV) that are then fused into a single HDR image~\cite{Debevec97,Mertens07}, with some of those methods including frame alignment~\cite{kalantari17, sen12, wu18} or pixel rejection strategies~\cite{yan19}.

The current state-of-the-art in multi-frame HDR fusion is significantly dominated by Convolutional Neural Networks (CNNs), especially for complex dynamic scenes~\cite{liu2021ntire, perezpellitero21, kalantari17, wu18, prabhakar20, yan19}. However, the computational requirements of most of the aforementioned state-of-the-art methods are far from real-time or even interactive processing on most hardware, especially when performing fine-grained motion estimation between frames. Despite several efforts in other computational photography applications~\cite{Maggioni2021,Ehmann2018}, HDR imaging has received limited attention in terms of computational efficiency.

The NTIRE 2022 HDR Challenge aims to stimulate research for efficient, low-complexity computational HDR imaging and better understand the state-of-the-art landscape of multiple frame HDR processing. It is part of a broad spectrum of associated challenges within the NTIRE 2022 workshop: spectral recovery~\cite{arad2022ntirerecovery}, 
spectral demosaicing~\cite{arad2022ntiredemosaicing},
perceptual image quality assessment~\cite{gu2022ntire},
inpainting~\cite{romero2022ntire},
night photography rendering~\cite{ershov2022ntire},
efficient super-resolution~\cite{li2022ntire},
learning the super-resolution space~\cite{lugmayr2022ntire},
super-resolution and quality enhancement of compressed video~\cite{yang2022ntire},
high dynamic range~\cite{perezpellitero2022ntire},
stereo super-resolution~\cite{wang2022ntire}, and burst super-resolution~\cite{bhat2022ntire}.

\section{Challenge}

The NTIRE 2022 HDR Challenge addresses the efficient HDR image enhancement task. This challenge aims to gauge and advance the state-of-the-art on HDR imaging, with a particular emphasis on efficient and low-complexity solutions. It focuses on challenging scenarios for multi-frame HDR image reconstruction, \ie wide range of scene illumination, accompanied by complex motions in terms of camera, scene and light sources. The remainder of this section describes the challenge set-up, including the dataset, as well as how the challenge tracks are designed and evaluated.

\subsection{Dataset}

For the NTIRE 2022 HDR Challenges, we follow the previous edition~\cite{perezpellitero21} data set-up, consisting of a curated dataset composed of approximately 1500 training, 60 validation and 201 testing examples. Each example in the dataset is in turn composed of three input LDR images, \ie\,short, medium and long exposures, and a related ground-truth HDR image aligned with the central medium frame. The images are collected from the work of Froelich \etal~\cite{Froehlich14}, where they capture an extensive set of diverse and challenging HDR videos using a professional two-camera rig with a semitransparent mirror. Input images are obtained using a pixel-measurement model~\cite{Hasinoff10}, which includes several sources of noise. For more details about the imaging and noise models, as well as other details, we refer the reader to \cite{perezpellitero21}.

\begin{table*}
\centering
\renewcommand{\arraystretch}{1.5}
\begin{tabular}{l c c c c c r} \toprule
    Team & Username & PSNR & \textbf{PSNR-$\mu$} & Runtime (s) & GMACs & Param.~$\times10^3$ \\ \midrule
    ALONG & Good & $39.417~_{(1)}$ & $37.424~_{(1)}$ & 0.324 & 198.47 & 489.01 \\
    antins\textunderscore cv & lishen & $38.607~_{(3)}$ & $37.252~_{(2)}$ & 0.185 & 198.38 & 576.23 \\
    XPixel-UM & Xy\textunderscore Chen & $38.015~_{(6)}$ & $37.209~_{(3)}$ &  0.276 & 199.88 & 1013.25\\ 
    AdeTeam & yeeah & $39.001~_{(2)}$	& $37.163~_{(4)}$ & 0.134 & 156.12 & 188.99 \\
    CZCV & pepper26545 & $37.388~_{(10)}$ & $36.972~_{(5)}$ & 0.431 & 193.93 & 633.69 \\
    BOE-IOT-AIBD & NBCS & $38.131~_{(5)}$ & $36.851~_{(6)}$ & 0.169 & 199.58 & 526.25\\
    VAlgo & buzzli & $37.888~_{(7)}$ & $36.723~_{(7)}$ & 0.061 & 175.25 & 1229.61\\
    MegHDR & liuzhen & $37.611~_{(9)}$ & $36.608~_{(8)}$ & 0.249 & 199.16 & 103.15 \\
    WorkFromHome & gaogao & $37.663~_{(8)}$ & $36.451~_{(9)}$ & 0.057 & 198.91 & 274.13\\
    TeamLiangJian & dth914 & $38.159~_{(4)}$ & $36.386~_{(10)}$ & 0.213 & 192.88 & 1628.27\\
    TVHDR & leochangliu & $37.221~_{(12)}$ & $36.089~_{(11)}$ & 0.573 & 199.02 & 753.52\\
    ForLight & FangyaLi & $37.370~_{(11)}$ & $35.988~_{(12)}$ & 0.226 &199.39 &80.65\\ 
    IMVIA & steventel & $36.185~_{(13)}$ & $34.588~_{(13)}$ & 0.045 & 118.99 & 38.83\\ 
    DGU-CILAB & cilab & $35.405~_{(14)}$ & $34.525~_{(14)}$ & 0.441 & 199.11 & 1300.60\\ 
    CVIP & hannah1258 & $37.028$\hspace{4ex} & $36.050$\hspace{4ex} & 0.8876 & 4818.98 & 2485.60 \\
    \textit{no processing} & - & $27.408$\hspace{4ex}  & $25.266\hspace{4ex} $ & - & - & - \\ 
    \bottomrule
\end{tabular}
\caption{Results and rankings of methods submitted to \textit{Track 1: Fidelity (low-complexity constraint)}. Participants are asked to optimize fidelity scores (PSNR and PSNR-$\mu$) while keeping complexity under $200$ GMACs. We denote in bold the main metric of the track.}
\label{table:track1-test}
\end{table*}

\begin{table*}
\centering
\renewcommand{\arraystretch}{1.5}
\begin{tabular}{l c c c c c r} \toprule
    Team & Username & PSNR  & PSNR-$\mu$ & Runtime (s) & \textbf{GMACs} & Param.~$\times10^3$ \\ \midrule
    ALONG & Good    & 38.843 & 37.033 & $0.183~_{(3)}$  & $74.02~_{(1)}$ & 177.41 \\
    antins\textunderscore cv & QianPaiChiGua & 38.766 & 37.100 & $0.155~_{(2)}$  & $146.28~_{(2)}$ & 393.73 \\
    AdeTeam & yeeah & 39.001 & 37.163 & $0.136~_{(1)}$  & $156.12~_{(3)}$ & 188.99 \\
    BOE-IOT-AIBD & NBCS & 38.150 & 37.225 & $0.473~_{(6)}$  & $756.78~_{(4)}$ & 1355.10 \\ 
    MegHDR & liuzhen & 38.749 & 37.030 & $0.269~_{(4)}$  & $790.48~_{(5)}$ & 401.83 \\
    0noise & Juan & 38.495 & 37.114 & $0.738~_{(7)}$  & $1769.85~_{(6)}$ & 1222.04 \\
    Winterfell & XuZiyao & 38.205 & 37.037 & $0.437~_{(5)}$  & $2073.42~_{(7)}$ & 1466.21\\
    ACALJJ32 & ACALJJ32 & 39.042 & 37.273 & $1.825~_{(8)}$  & $5704.17~_{(8)}$ & 8218.37 \\ 
    \emph{Baseline} (AHDR~\cite{yan19}) & - & 37.597 & 37.021 & 0.760\hspace{3ex} & $2916.92$\hspace{3ex} & 1441.28\\
    TeamLiangJian & dth914 & 38.587 & 36.858 & 0.820\hspace{3ex} & $2260.30$\hspace{3ex}  & 20476.96\\
    KCML2 & kcml2 & 38.969 & 36.302 & 1.110\hspace{3ex} & $6000.00$\hspace{3ex} & 2499.00 \\
    \textit{no processing} & - & $27.408$  & $25.266 $ & - & - & - \\ 
    \bottomrule
\end{tabular}
\caption{Results and rankings of methods submitted to the \textit{Track 2: Low-Complexity (fidelity constraint)}. Participants are asked to minimize the complexity of their solutions (GMACs and runtime) while achieving at least the same fidelity scores (PSNR and PSNR-$\mu$) as the baseline method AHDR~\cite{yan19}. We denote in bold the main metric of the track.}
\label{table:track2-test}
\end{table*}

\subsection{Challenge Design and Tracks}
This challenge focuses on developing efficient multi-frame HDR reconstruction methods. The challenge is organized in two different tracks that focus on different aspects of the fidelity-complexity trade-off.

\subsubsection{Track 1: Fidelity (low-complexity constraint)}
This track enforces a strict low-complexity constraint, requiring all submitted methods to be less than $200$ GMACs for a testing input tensor of size $(3, 3, 1900, 1060)$, \ie $($sequence length, channels, width, height$)$. All methods are required to perform any pre- or post-processing within the model definition, \eg gamma correction, normalization, exposure alignment, downsampling. The submitted methods are then ranked based on fidelity metrics, with an focus on the tonemapped PSNR-$\mu$. Participants are thus invited to optimize the fidelity scores while ensuring they meet the constraint on computational complexity. 

\subsubsection{Track 2: Low Complexity (fidelity constraint)}
This track enforces a fidelity constraint, requiring all submitted methods to achieve greater PSNR and PSNR-$\mu$ values than the baseline model, AHDR \cite{yan19}. The submitted methods are then ranked based on complexity metrics, with an emphasis on the number of GMACs. As in Track 1, GMACs are calculated with an input tensor of size $(3, 3, 1900, 1060)$. Participants are thus invited to minimize the computational complexity, while ensuring they meet the constraint on fidelity.

\subsection{Challenge Phases}
The challenge consists of two distinct phases, a development phase intended to allow participants to improve and validate their models and a testing phase designed to evaluate final submissions.

\subsubsection{Development Phase}
Participants are given access to the complete training set, including the input LDR sequences and corresponding HDR ground truths. In addition, participants have access to only the input LDR sequences of the validation set. Each track respective constraints are announced (complexity limit for Track 1 and baseline performance on the validation set for Track 2). The participants are able to compute fidelity metrics on the validation set by uploading their predictions to the Codalab challenge server, and are given access to scripts to compute complexity metrics which are in turn self-reported in the Codalab as well. The validation leaderboard is visible to all participants.

\subsubsection{Testing Phase}
Participants gain access to the input sequences of the testing set and are then asked to submit a subset of their HDR predictions ($67$ images) to the Codalab evaluation server and e-mail the complete set ($201$ images), code and factsheet to the organizers. Unlike the development phase, the results and related leaderboards are not visible to the participants during the testing phase. Once the testing phase is over (\ie after 7 days), the organizers verify and run the provided code to obtain the final results.

\subsection{Evaluation}
\subsubsection{Fidelity Metrics}
The evaluation of the challenge submissions is based on the computation of objective metrics between the estimated and the ground-truth HDR images. We use the well-known standard peak signal-to-noise ratio (PSNR) both directly on the estimated HDR image and after applying the $\mu$-law tonemapper, which is a simple and canonical operator used widely for benchmarking in the HDR literature \cite{kalantari17,prabhakar20,yan19}. Within these two metrics, we selected PSNR-$\mu$ as the primary metric to rank methods in the challenge.

For the PSNR directly computed on the estimated HDR images, we normalize the values to be in the range $[0,1]$ using the peak value of the ground-truth HDR image. 

For the PSNR-$\mu$, we apply the following tone-mapping operation $\mathrm{T}(H)$:
\begin{equation}
\label{eq:mu-law}
\mathrm{T}(H)  = \frac{\log(1 + \mu H)}{\log(1 + \mu)}
\end{equation}
where $H$ is the HDR image, and $\mu$ is a parameter that controls the compression, which we fix to $\mu = 5000$ following common HDR evaluation practices. In order to avoid excessive compression due to peak value normalization, for the PSNR-$\mu$ computation, we normalize using the $99^{\text{th}}$ percentile of the ground-truth image followed by a $\tanh$ function to maintain the $\left[0,1\right]$ range.
\subsubsection{Complexity Metrics}
To assess the complexity of the submitted solutions, we measure the number of required operations in terms of Multiply-Accumulate operations (MAC) and runtime (in seconds). Both metrics are self-reported by participants during the development phase but measured by the organizers during the final testing phase using a unified hardware set-up, namely an Intel Xeon Platinum 8168 CPU (2.70GHz) and a single NVIDIA Tesla V100 PCIe 32 GB. We choose GMACs as the primary metric relating to low complexity for the challenges as this is independent of the platform used and thus can provide meaningful ranking during the development phase. However, we incorporate both metrics in assessing the solutions complexity in the final testing phase. For completeness, we also measured the number of parameters this metric is however not directly used for the challenge.

\section{Results}
\begin{figure*}
\includegraphics[width=1.0\textwidth]{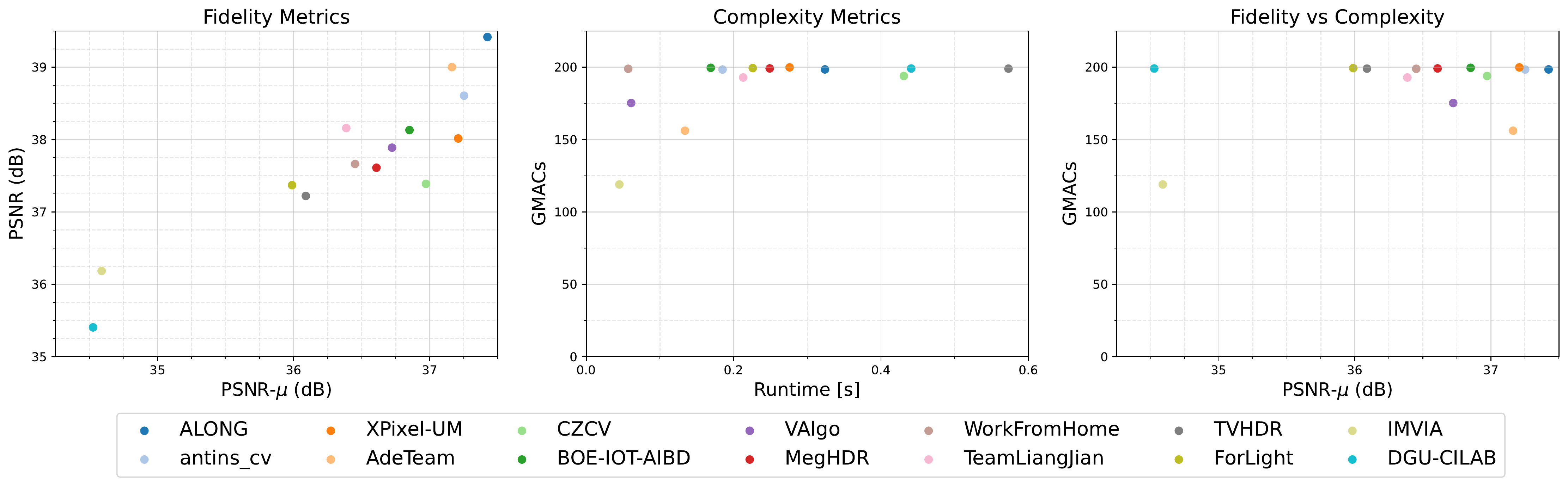}
\caption{Several visualizations of the challenge \textit{Track 1} methods, from left to right: combined fidelity metrics (PSNR-$\mu$ and PSNR), combined complexity metrics (GMACs and runtime) and combined fidelity and complexity metrics (GMACs and PSNR-$\mu$).}
\label{fig:track1} 
\end{figure*}

\begin{figure*}
\includegraphics[width=1.0\textwidth]{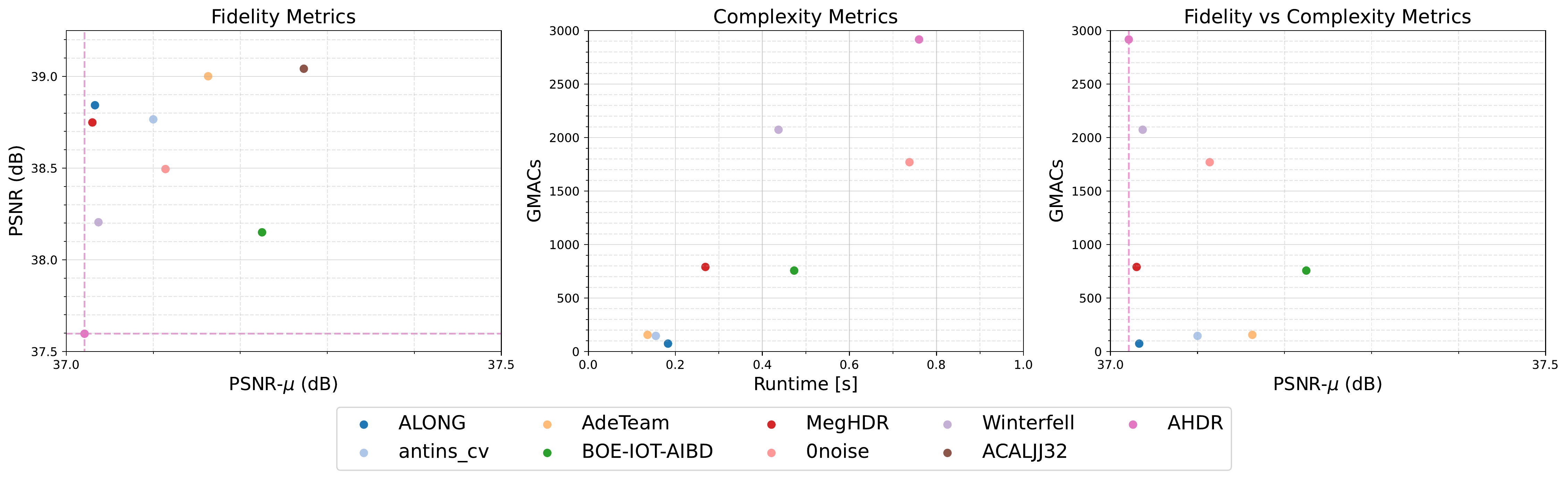}
\caption{Several visualizations of the challenge \textit{Track 2} methods, from left to right: combined fidelity metrics (PSNR-$\mu$ and PSNR), combined complexity metrics (GMACs and runtime) and combined fidelity and complexity metrics (GMACs and PSNR-$\mu$).}\label{fig:track2} 
\end{figure*}
From 197 registered participants in Track 1, 15 teams entered the final testing phase and submitted results, source code and fact sheets. All submissions but one complied with the eligibility criteria of under $200$ GMACs constraint. As for Track 2, from 168 registered participants, 10 teams entered the final testing phase and submitted results, source code and fact sheets. In this track, two teams were not able to reach the minimal PSNR scores as set by the AHDR baseline~\cite{yan19} and thus are not eligible. We report the final test phase results in Table~\ref{table:track1-test} and \ref{table:track2-test} for Track 1 and 2 respectively. We show more visualizations of fidelity and complexity metrics for each track in Figure~\ref{fig:track1} and \ref{fig:track2}.

The methods and the teams that entered the final phase are described in Section~\ref{sec:team-and-methods}, more detailed information about each team and their members' affiliation can be found in Appendix~\ref{ap:teams-and-affiliations}.

\subsection{Main ideas}

The emphasis on low complexity and efficiency forced participants to aggressively reduce GMACs compared to last year's challenges~\cite{perezpellitero21}, while maintaining or even improving performance. One dominant trend is the use of depthwise and separable convolutions to reduce computational complexity. This strategy was employed by many of the top performers. The use of residual feature distillation blocks (RFDBs) is also seen as an efficient replacement for the dilated residual dense blocks (DRDB) backbone. Another trend that led to increased performance is the use of vision transformer architectures, either directly as part of the model backbone or as a teacher for knowledge distillation. A large majority of participants use a U-Net architecture due to its natural efficiency and performance advantages, and choose to process input images in feature space at a downsampled resolution or at multiple scales. Participants also use pixel shuffle layers to reduce computational complexity, and we note that participants typically trained models for more epochs when compared to the NTIRE 2021 HDR Challenge edition. Finally, the use of spatial attention, as introduced by AHDR\cite{yan19}, is practically ubiquitous across the different architectures.

\subsection{Top results}

\textbf{Track 1}: The top three methods are: ALONG, antins\_cv and XPixel-UM. There is a moderate difference of $0.17$ dB in PSNR-$\mu$ between Along and antins\_cv, and a small difference of $0.04$ dB between antins\_cv and XPixel-UM. There is a much more significant difference in PSNR, with $0.81$ dB between ALONG and antins\_cv, and a further $0.59$ dB between antins\_cv and XPixel-UM. As expected, all methods are close to but just under $200$ GMACs. Also of notable mention is the method submitted by AdeTeam which ranks fourth in PSNR-$\mu$ (by only $0.046$ dB) but ranks second in PSNR and has a faster runtime and significantly fewer parameters than the top three solutions. Most of the remaining methods achieve scores of between $0.4$-$1.4$ dB lower than the first-ranking solution.

\textbf{Track 2}: The top three methods are: ALONG, antins\_cv and AdeTeam. ALONG achieves the required fidelity constraint using only $74.02$ GMACs, while antins\_cv and AdeTeam require $146.28$ and $156.12$ GMACs respectively. There is a large difference between these and the remaining methods, which use between $\times10$ and $\times77$ as many GMACs as the first-ranked solution. We note that the first-ranked solution is able to achieve a $\times39$ reduction in GMACs compared to the AHDR baseline while maintaining the same performance.

\section{Team and Methods}
\label{sec:team-and-methods}
\subsection{ALONG}
\begin{figure}[h]
\centering
\includegraphics[width=1.0\columnwidth]{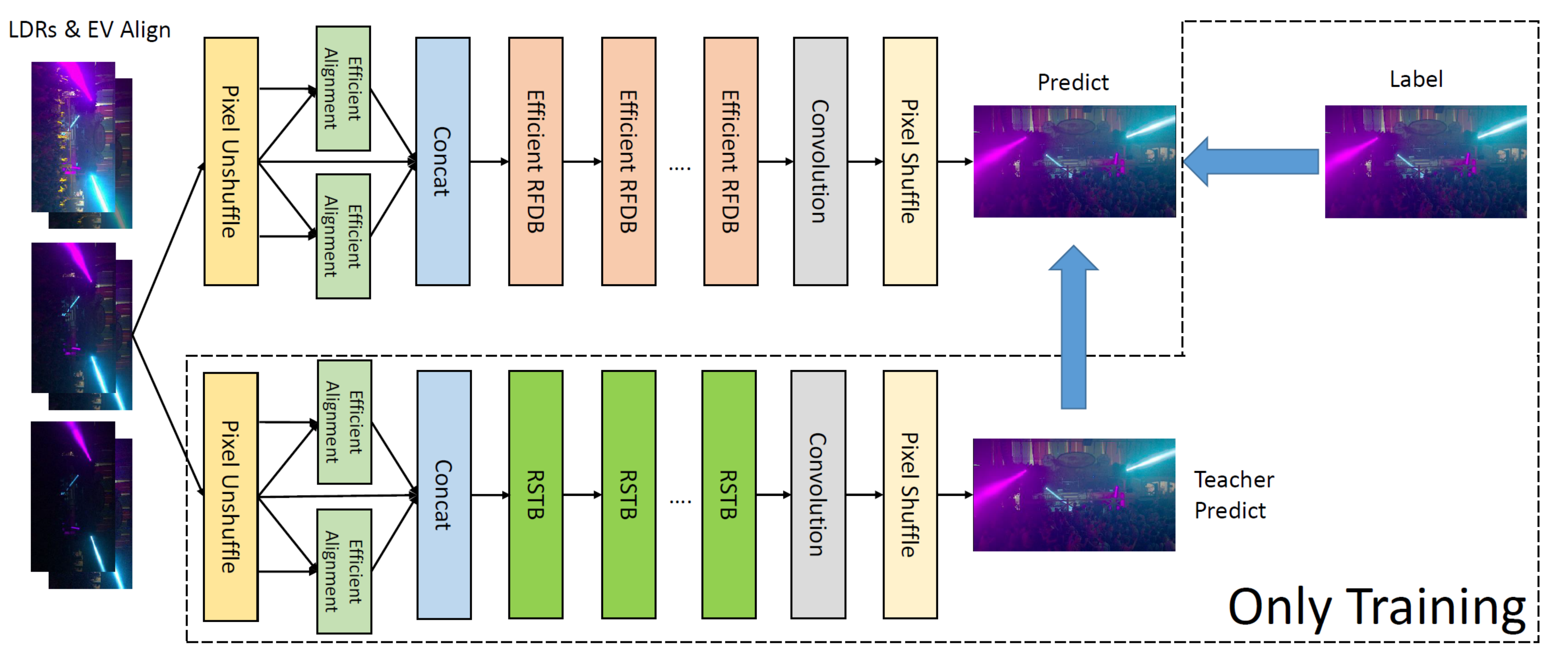}
\caption{Architecture of EfficientHDR: Residual Feature Network and Transformer Teacher for Efficient High Dynamic Range, proposed by the ALONG team.}
\label{fig:along} 
\end{figure}

The team presents a solution that is composed of attention alignment, feature extraction and reconstruction stages, with an emphasis on efficient architecture backbones and teacher-student distillation training. Figure~\ref{fig:along} shows an overview of the proposed approach.
Firstly, 6 images (3 LDR images and 3 exposure aligned images) are fed to a pixel shuffle layer, reducing the feature spatial resolution. An efficient alignment module follows the attention module of AHDRNet~\cite{yan19}. The feature extraction uses an RFDB~\cite{Liu20a} backbone with the ESA block in RFANet~\cite{Liu20b} rather than other common architectures \eg DRDB~\cite{yan19}. Depth-wise and point-wise convolutions replace the conventional convolutions in both RFDB and the attention alignment module. Additionally, the ReLU activation is substituted by the Sigmoid Linear Unit (SiLU)~\cite{lin2022revisiting}. Finally, the solution adopts a knowledge-distillation training strategy, where the teacher's output images and ground-truth images are used to guide the student network.

\subsection{Antins\textunderscore cv}
\begin{figure}[h]
\centering
\includegraphics[width=1.0\columnwidth]{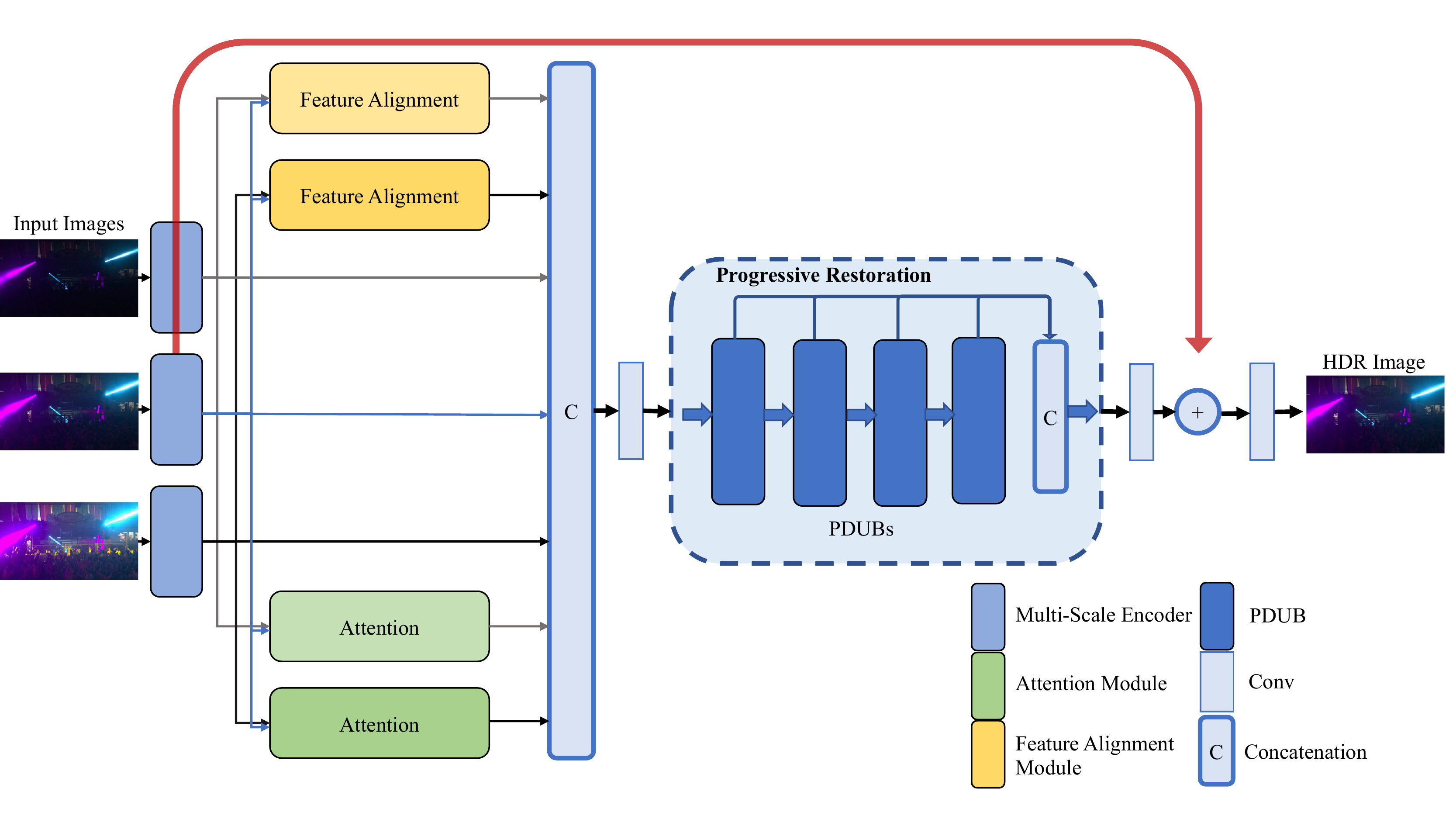}
\caption{Architecture of Efficient Progressive High Dynamic Range Image Restoration via
Attention and Alignment Network, proposed by the antins\textunderscore cv team.}
\label{fig:antinscv} 
\end{figure}

The team proposes an efficient progressive HDR network~\cite{gaochengyu2022ntire} based on AHDRNet\cite{yan19} and ADNet\cite{liu21_ntire}.
Figure~\ref{fig:antinscv} shows the overall architecture.
Firstly, considering noise, a multi-scale encoder layer is proposed to extract high and low-frequency features from the input signals.
Secondly, to deal with misalignment problems, following~\cite{young2022feature}, a feature-alignment module is introduced in place of the deformable convolutional network~\cite{dai2017deformable}, leading lower computational cost. Finally, a Progressive Dilated U-shape Block (PDUB) is proposed, composed of multiple tiny u-form structures to progressively restore features, achieving a good balance between PSNR-$\mu$ and GMACs in both Track 1 and Track 2. Compared to DRDB\cite{yan19}, the proposed PDUB does not work at full resolution and supports a progressive plug-and-play mode to support models with different GMACs requirements. Separable convolution~\cite{related_mbnv1, related_mbnv2, related_mbnv3} is applied throughout the main body of the network yielding improved low-complexity performance. 

\subsection{XPixel-UM}
\begin{figure}[h]
\centering
\includegraphics[width=1.0\columnwidth]{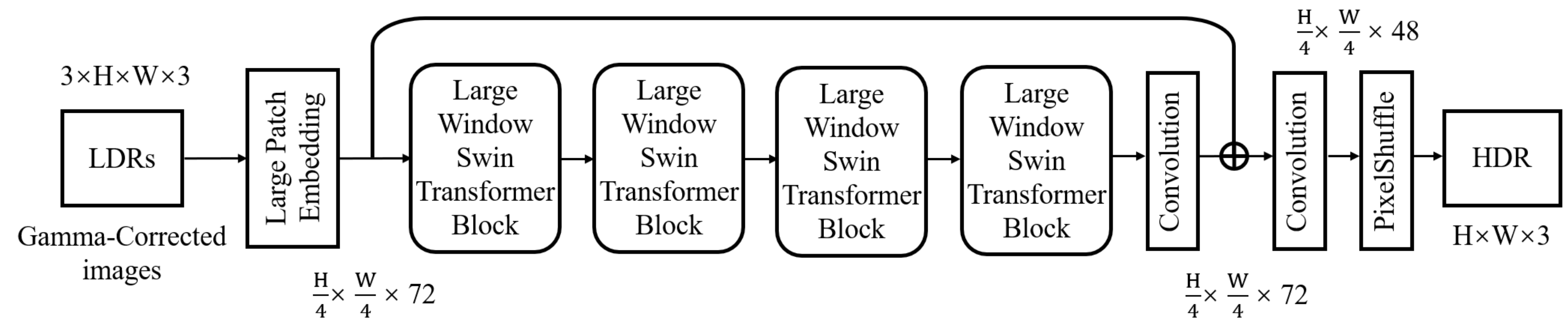}
\caption{Architecture of SwinHDR: Efficient Swin Transformer for High Dynamic Range Imaging, proposed by the XPixel-UM team.}
\label{fig:xpixel-um} 
\end{figure}

The team introduces a Transformer-based method for HDR imaging based on the Swin Transformer~\cite{liu2021swin}. As shown in Figure~\ref{fig:xpixel-um}, the overall architecture of this approach mainly consists of a large patch embedding module, a series of Swin Transformer blocks and a pixel shuffle layer. Specifically, a convolutional layer with kernel size and stride of 4 is utilized for the large patch embedding to map the concatenated input LDRs to a high-dimensional representation. Then, four Swin Transformer blocks similar to~\cite{liang2021swinir} with a window size of 16 are adopted to extract the deep features. A long skip connection is added for better optimization, and a pixel shuffle layer is used to reconstruct the high-resolution HDR result. 

\subsection{AdeTeam}
\begin{figure}[h]
\centering
\includegraphics[width=1.0\columnwidth]{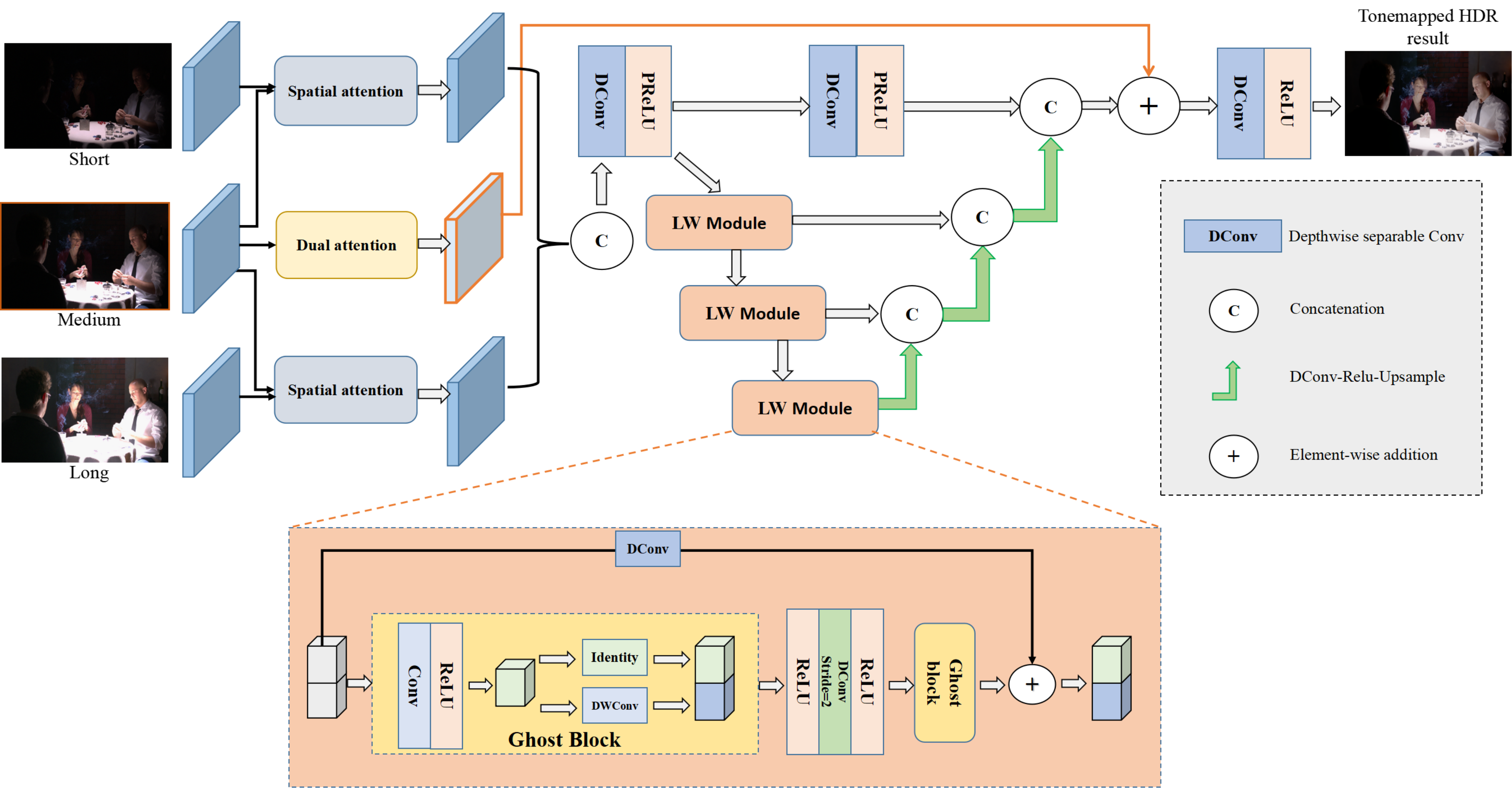}
\caption{Architecture of A Lightweight HDR Deghosting Network with LW Module, proposed by AdeTeam.}
\label{fig:adeteam} 
\end{figure}

The team presents a CNN with both high accuracy and high efficiency for reconstructing ghosting-free HDR images~\cite{yan2022ntire}. As shown in Figure~\ref{fig:adeteam}, a hybrid framework is proposed to fuse high-resolution features and multi-scale features with a lightweight block. Unlike previous CNN-based methods, the high-resolution and encoder-decoder structure are integrated into a model. Following AHDRNet \cite{yan19}, spatial attention is used to remove motion and refine the features of non-reference images during the feature extraction stage. Additionally, a residual operation, which highlights useful information based on the dual attention module, forces the fusion stage to learn more details about degenerated regions. In the fusion stage, inspired by \cite{han2020ghostnet}, a lightweight (LW) module fuses features with fewer parameters. For the high-resolution branch, depthwise separable convolution~\cite{chollet2017xception} is used to maintain high-resolution features. On the other hand, since the encoder-decoder network tends to rapidly capture a larger receptive field for HDR deghosting, the LW block is inserted into the encoder-decoder network to learn different scale features. Finally, the estimated image is generated with a depthwise separable convolution.

\subsection{BOE-IOT-AIBD}
\begin{figure}[h]
\centering
\includegraphics[width=1.0\columnwidth]{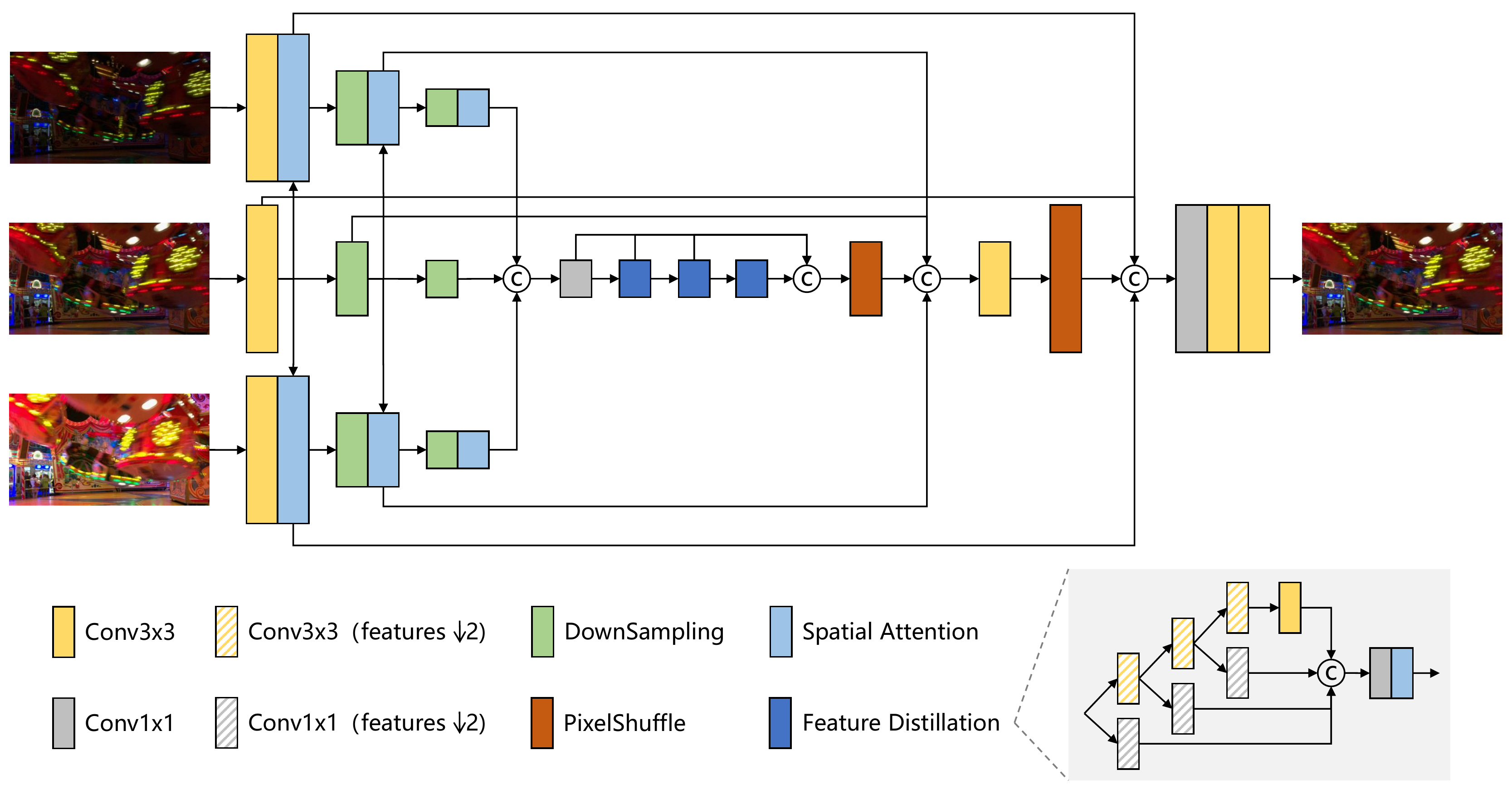}
\caption{Architecture of Multi-Level Attention U-Net for HDR Reconstruction,
proposed by the BOE-IOT-AIBD team.}
\label{fig:BOE-IOT-AIBD} 
\end{figure}

The team proposes a multi-level attention U-Net architecture. The overall architecture used for Track 1 is shown in Figure~\ref{fig:BOE-IOT-AIBD}. The team adopts spatial attention modules \cite{liu21_ntire, Liu20b} to select the most appropriate regions of LDR images and corresponding exposure aligned images for feature fusion. A multi-branch U-Net~\cite{unet} architecture is used to process and merge features at three different resolutions. At the low resolution, image features are concatenated and processed by feature distillation modules \cite{Liu20a} in which feature channels are halved gradually. A $1\times1$ convolutional layer is used after each feature concatenation, and downsampling is implemented by the a combination of pixel shuffle and two convolutional layers of kernel size 1 and 3 respectively.

In Track 2, the framework is the same as that submitted to Track 1. To further increase the network performance, the team utilizes spatial attention and the feature fusion group (FFG)~\cite{muqeet2020multi} that consists of several multi-attention blocks (MAB) to replace the feature distillation. The MAB makes the attention mechanisms more efficient by introducing dilated convolutions with different filter sizes.

\subsection{CZCV}
\begin{figure}[h]
\centering
\includegraphics[width=1.0\columnwidth]{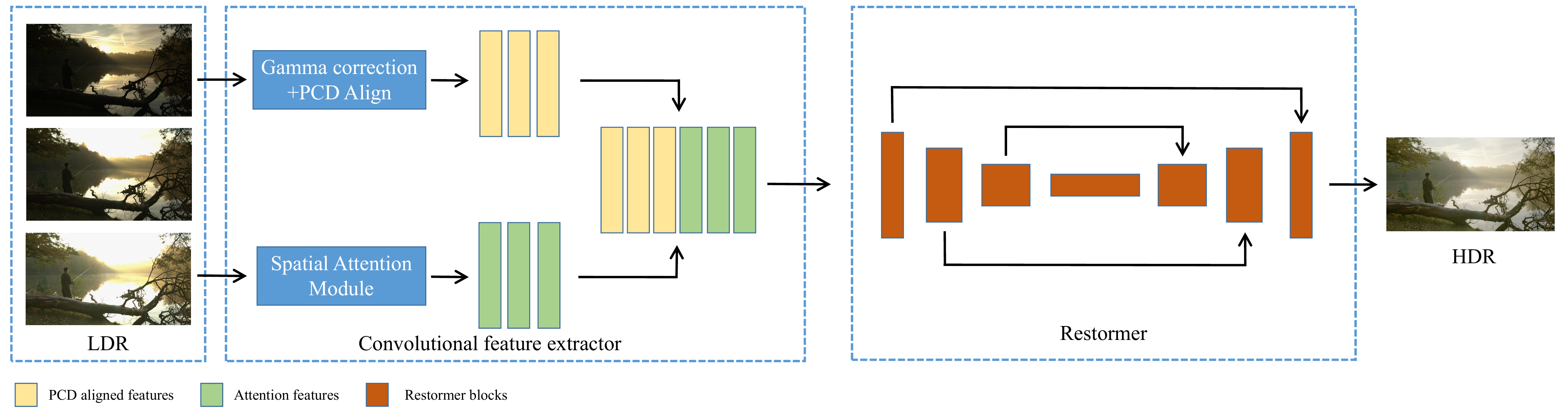}
\caption{Architecture of Multi-frame High Dynamic Range Image Reconstruction using Transformer, proposed by the CZCV team.}
\label{fig:CZCV_network} 
\end{figure}

The team exploits the transformer architecture for HDR reconstruction. The proposed solution is presented in Figure~\ref{fig:CZCV_network}. It consists of a convolutional feature extractor for feature extraction and a transformer for HDR reconstruction. The team builds the convolutional feature extractor using part of ADNet~\cite{liu21_ntire}, including exposure alignment, the pyramid, cascading and deformable (PCD) alignment module~\cite{wang19}, and the spatial attention module. The transformer part employs the recently proposed Restormer architecture~\cite{Zamir22Restormer}, due to its reported low complexity and high performance. To further reduce the complexity of the model, the team replaces some regular convolution layers with depth-wise convolution followed by $1 \times 1$ convolution.

\subsection{MegHDR}
\begin{figure}[h]
	\centering
	\includegraphics[width=1.0\columnwidth]{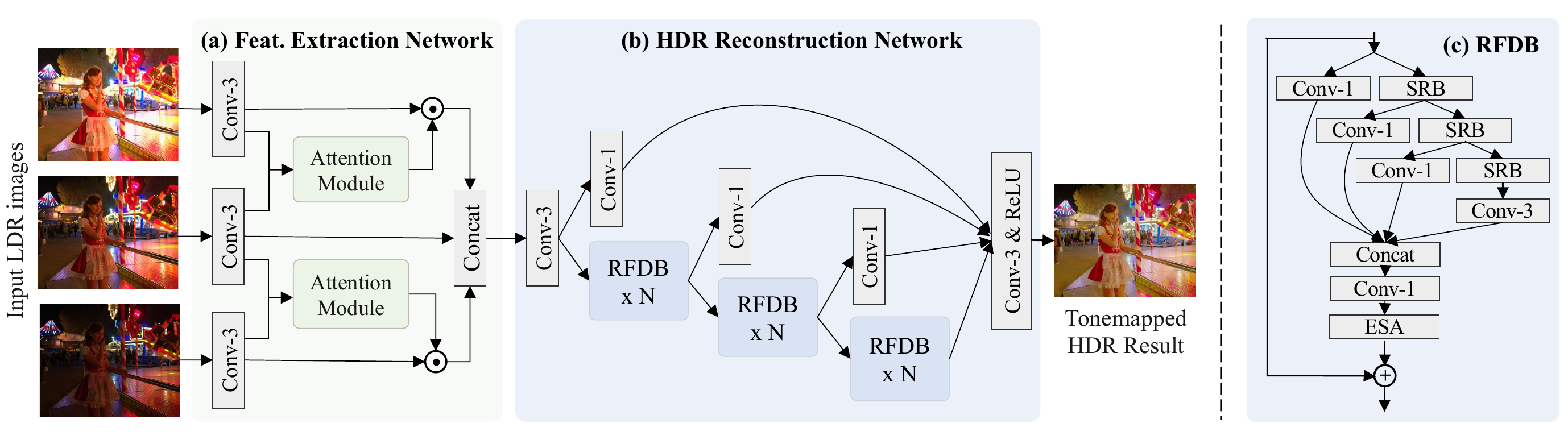}
	\caption{Architecture of Multi-stage Feature Distillation Network for Efficient High Dynamic Range Imaging, proposed by the MegHDR team.}
	\label{fig:MegHDR} 
\end{figure}

For efficient HDR imaging, the team proposes a multi-stage feature distillation network (MFDN). The architecture of the proposed network is illustrated in Figure~\ref{fig:MegHDR}. The proposed MFDN mainly consists of two parts: 1) the feature extraction network and 2) the HDR reconstruction network. Given the input LDR images, spatial features are extracted through a feature attention module. The feature extraction network follows previous work~\cite{yan19}, but in Track 1, the attention module is implemented using coordinate attention~\cite{hou21} to reduce the complexity of the model further. The extracted features are then fed into the HDR reconstruction network, reconstructing the corresponding HDR image. The reconstruction network consists of $N$ successive residual feature distillation blocks (RFDBs). The RFDBs are adopted to distil the coarse features (with noisy, under- and over-exposed regions), and the retained features in each stage are used to reconstruct noise-free high-quality HDR results. In Track 1, $N$ is set to 3 and the channel number of each RFDB is set to 16, whereas in Track 2, $N$ is set to 2 and the RFDB channels is set to 32 (refer to Figure~\ref{fig:MegHDR} for related parameter details).

\subsection{0noise}
\begin{figure}[h]
\centering
\includegraphics[width=1.0\columnwidth]{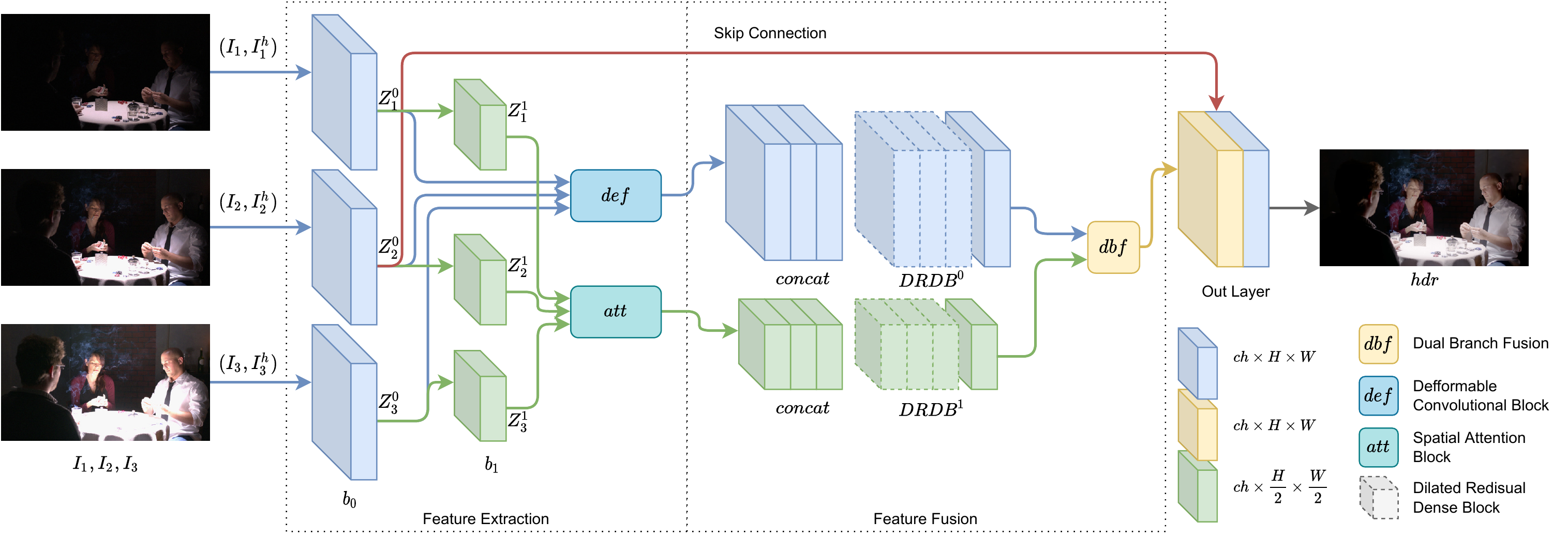}
\caption{Architecture of Dual Residual HDR (DRHDR) Network, proposed by the 0noise team.}
\label{fig:0noise} 
\end{figure}

The team presents Dual Branch Residual Network for HDR Imaging (DRHDR)~\cite{marinyega2022ntire}. The overall architecture is shown in Figure~\ref{fig:0noise} and comprises two routes, one branch that operates at full resolution, and a second branch that operates at a  fourth of the original resolution. 
The full resolution branch adopts a deformable convolutional block~\cite{liu21_ntire} while the low resolution branch adopts spatial attention~\cite{yan19}. Both branches utilize DRDBs. Features from both branches are fused through a Dual Branch Fusion. The output layer incorporates a dense global skip connection.

\subsection{VAlgo}
\begin{figure}[h]
	\centering
	\includegraphics[width=1.0\columnwidth]{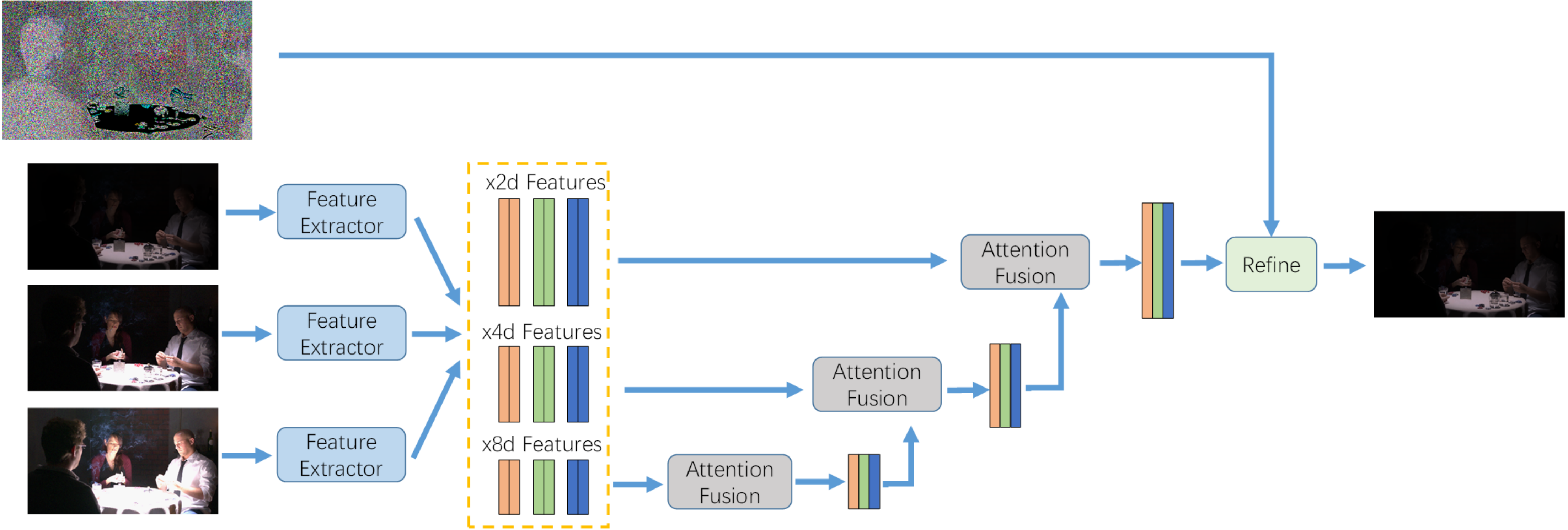}
	\caption{Architecture of AUNet for HDR, proposed by the VAlgo team.}
	\label{fig:aunet} 
\end{figure}

The team proposes a U-Net-style network with a spatial attention mechanism (AUNet). The overall architecture is shown in Figure~\ref{fig:aunet}. In order to fuse information from short and long exposure inputs, the team refers to AHDRNet\cite{yan19} and uses a spatial attention module to evaluate the importance of short and long images. To lower the computational complexity, inputs are downscaled to half of the original resolution and residual maps are saved to recover the final results. In the final stage, features and residual maps are concatenated to generate the final HDR image.

\subsection{Winterfell}
\begin{figure}[h]
	\centering
	\includegraphics[width=1.0\columnwidth]{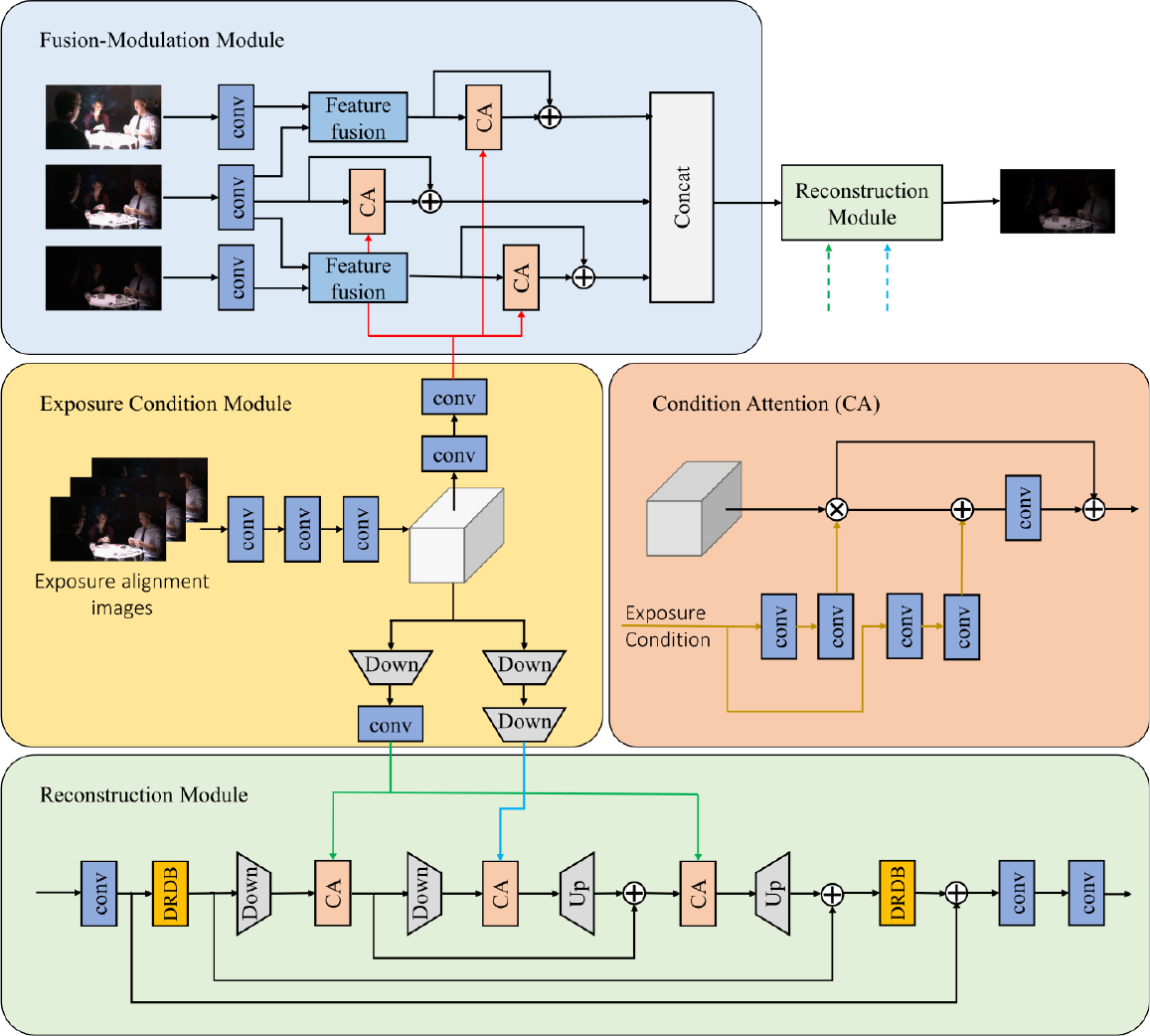}
	\caption{Architecture of Condition Attention-guided Reconstruction Network with Multi-exposure Fusion for High Dynamic Range Images, proposed by the Winterfell team.}
	\label{fig:winterfell} 
\end{figure}

The team propose a lightweight condition attention-guided reconstruction network (CARN) with multi-exposure fusion for HDR. The architecture of the proposed network is illustrated in Figure~\ref{fig:winterfell}. First, an exposure condition module generates the condition attention maps for the entire network. The structure of the exposure condition module is derived from~\cite{chen2021hdrunet}, and the input of this module involves the short, medium and long exposure alignment images. Meanwhile, the long and medium features are fused. Then, inspired by~\cite{wang18}, condition attention (CA) is introduced to adjust these fused features with exposure condition maps. The same procedure processes the short exposed image. Hence, long-medium fusion-modulation maps and short-medium fusion-modulation maps are obtained. The medium exposed features are also sent to a CA to generate self-fusion-modulation maps.
Finally, the concatenated fusion-modulation maps are fed to a reconstruction module to obtain the 16-bit HDR image. Specifically, the backbone of the reconstruction module is a variant of U-Net~\cite{Ronneberger2015UNet}. Dilated residual dense blocks (DRDBs) are adopted to distil the high-frequency information at a multi-scale level, and the condition attention maps are utilized at a lower scale to optimize the reconstruction maps.

\subsection{ACALJJ32}
\begin{figure}[h]
\centering
\includegraphics[width=1.0\columnwidth]{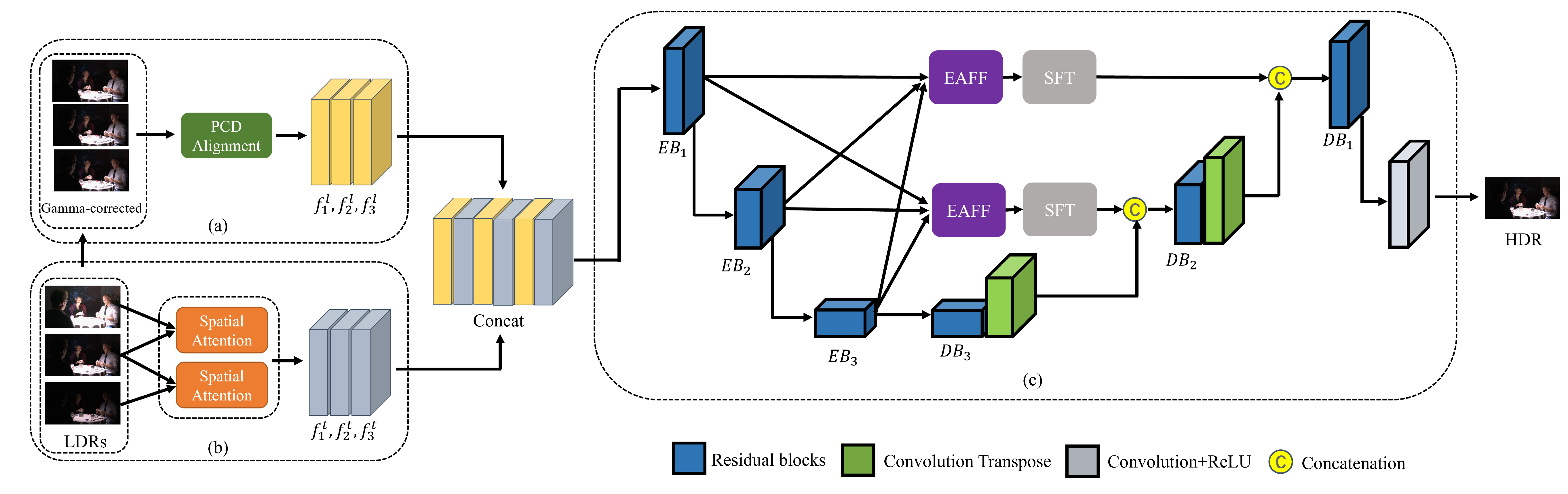}
\caption{Architecture of Multi-scale Asymmetric Learning for High Dynamic Range Imaging, proposed by the ACALJJ32 team.}
\label{fig:ma_unet} 
\end{figure}

The team presents a solution called MA-UNet to handle artifacts and colour aliasing in HDR image reconstruction. The network structure of MA-UNet is presented in Figure~\ref{fig:ma_unet}. First, the gamma-corrected LDR images are aligned by a Pyramid, Cascading and Deformable (PCD) alignment module~\cite{wang19}, and then the LDR images are fused by spatial attention modules~\cite{liu21_ntire}. Second, a U-Net structure is adopted to process the features into different scales. Instead of fusing the multi-scale features in a gradually increasing way, the proposed EAFF module~\cite{cho2021rethinking} connects features in parallel, which can remove the colour aliasing and hallucinate reasonable details. Third, to allow useful information to propagate further, the SFT module is proposed to process the outputs from the EAFF module. To reduce computational complexity, simple concatenation is used in EAFF and SFT is not adopted during the testing phase.

\subsection{WorkFromHome}
\begin{figure}[h]
\centering
\includegraphics[width=1.0\columnwidth]{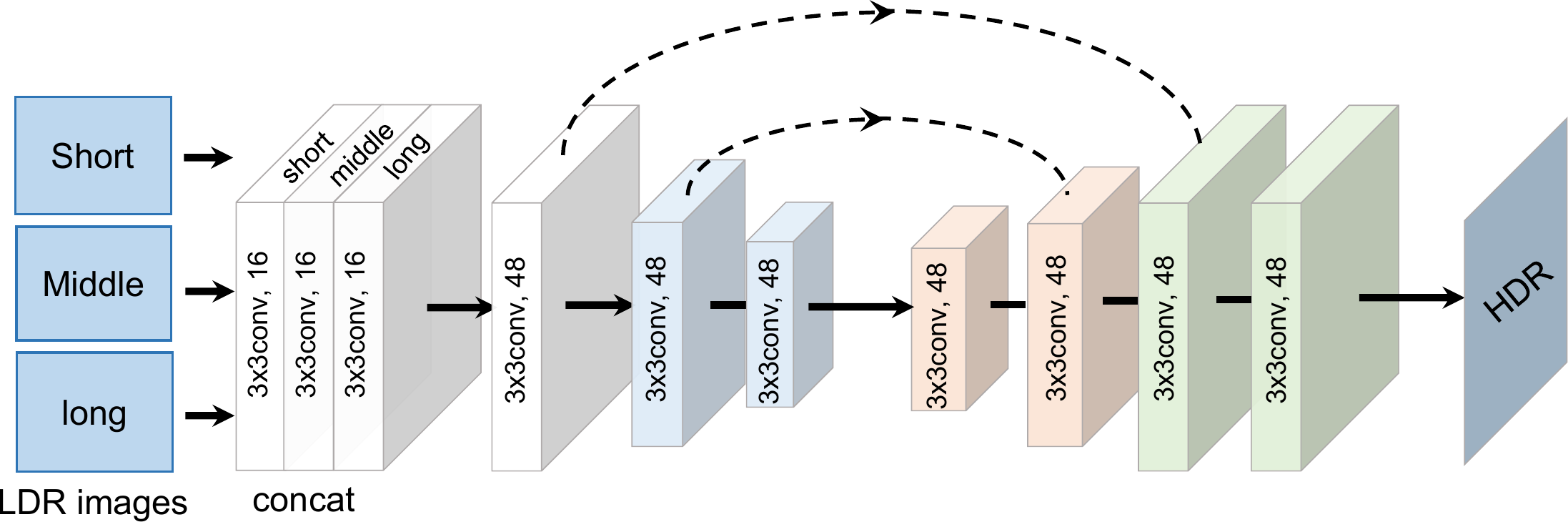}
\caption{Architecture of Deep Residual U-Net for High Dynamic Range Imaging, proposed by the WorkFromHome team.}
\label{fig:WFH} 
\end{figure}

The team proposes Deep Residual U-Net for High Dynamic Range Imaging. The model architecture, shown in Figure~\ref{fig:WFH}, takes three LDR images as input and directly outputs the synthesized HDR image. First, three convolutional kernels extract the visual feature maps of the three different LDR inputs and then concatenate them along the channel dimension. Then these features are fed into a U-Net architecture with 2 downsampling and 2 upsampling operations. Note, the dashed shortcuts in Figure~\ref{fig:WFH} are adding operations rather than concatenation operations used in regular U-Net. During this process, the image features of different exposure LDR inputs are fused thoroughly and finally transformed into the final HDR image with a regular convolutional kernel.

\subsection{TeamLiangJian}
\begin{figure}[h]
\centering
\includegraphics[width=1.0\columnwidth]{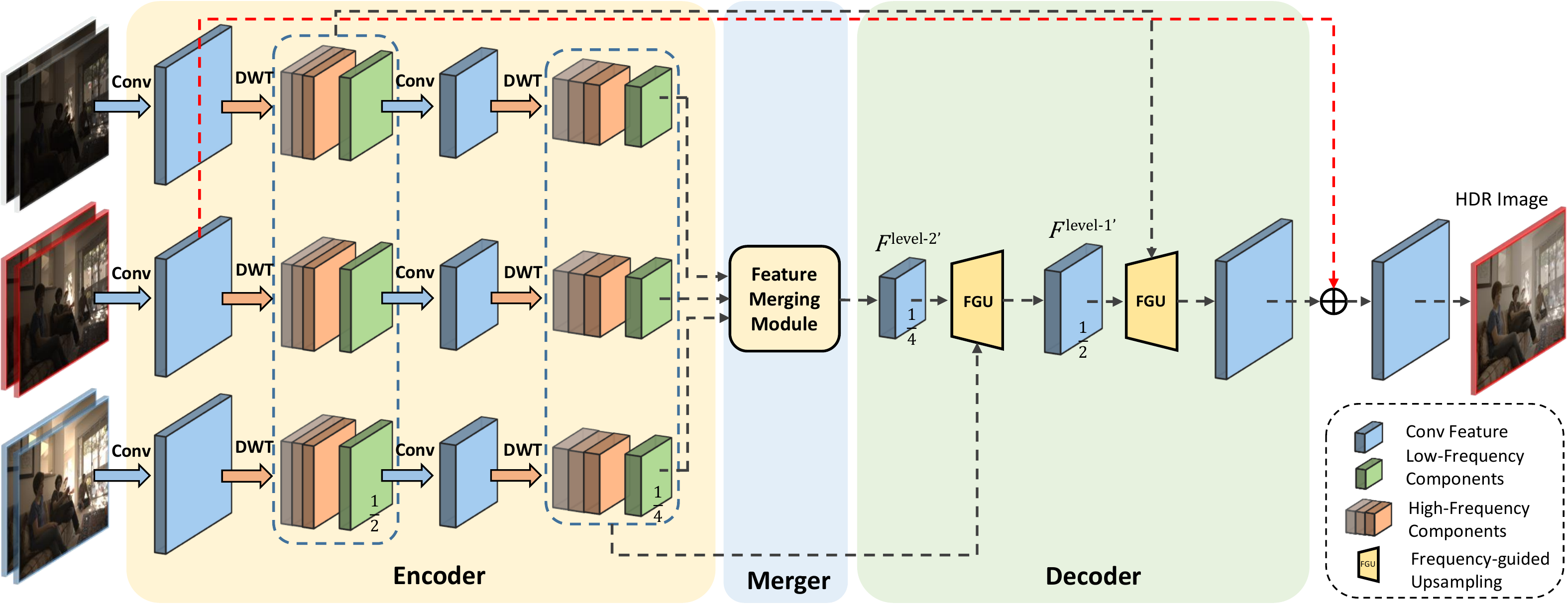}
\caption{Architecture of Wavelet-based Neural Network for Efficient High Dynamic Range Imaging, proposed by the TeamLiangJian team.}
\label{fig:liangjian} 
\end{figure}

The team presents a frequency-guided network (FHDRNet) to explicitly deal with signals of different frequency sub-bands for HDR imaging (see Figure~\ref{fig:liangjian}), using the discrete wavelet transform (DWT) to decompose the input into different frequency sub-bands. Specifically, ghosting artifacts are removed in the low-frequency sub-band, and high-frequency sub-bands are used to preserve details. The proposed network contains three parts: an encoder, a merger, and a decoder. In the encoder, the inputs 
are sent into three independent sub-networks. In each sub-network, the DWT is used for decomposing the feature maps into different frequency sub-bands, 
among which only the low-frequency sub-band 
is used for the next stage (scale) processing. All frequency sub-bands are also sent to the corresponding frequency-guided upsampling modules through skip connections. The merger fuses the three inputs (in the low-frequency sub-band) into a ghost-free one, which is then sent to the decoder. The merging module takes only low-frequency components from the previous stage as input and generates a merged result, focusing on structural information. In the decoder, the frequency-guided upsampling module is used to process features in the low-frequency and high-frequency sub-bands independently and then reconstruct the feature maps to a finer scale using the IDWT.

\subsection{TVHDR}
\begin{figure}[h]
\centering
\includegraphics[width=1.0\columnwidth]{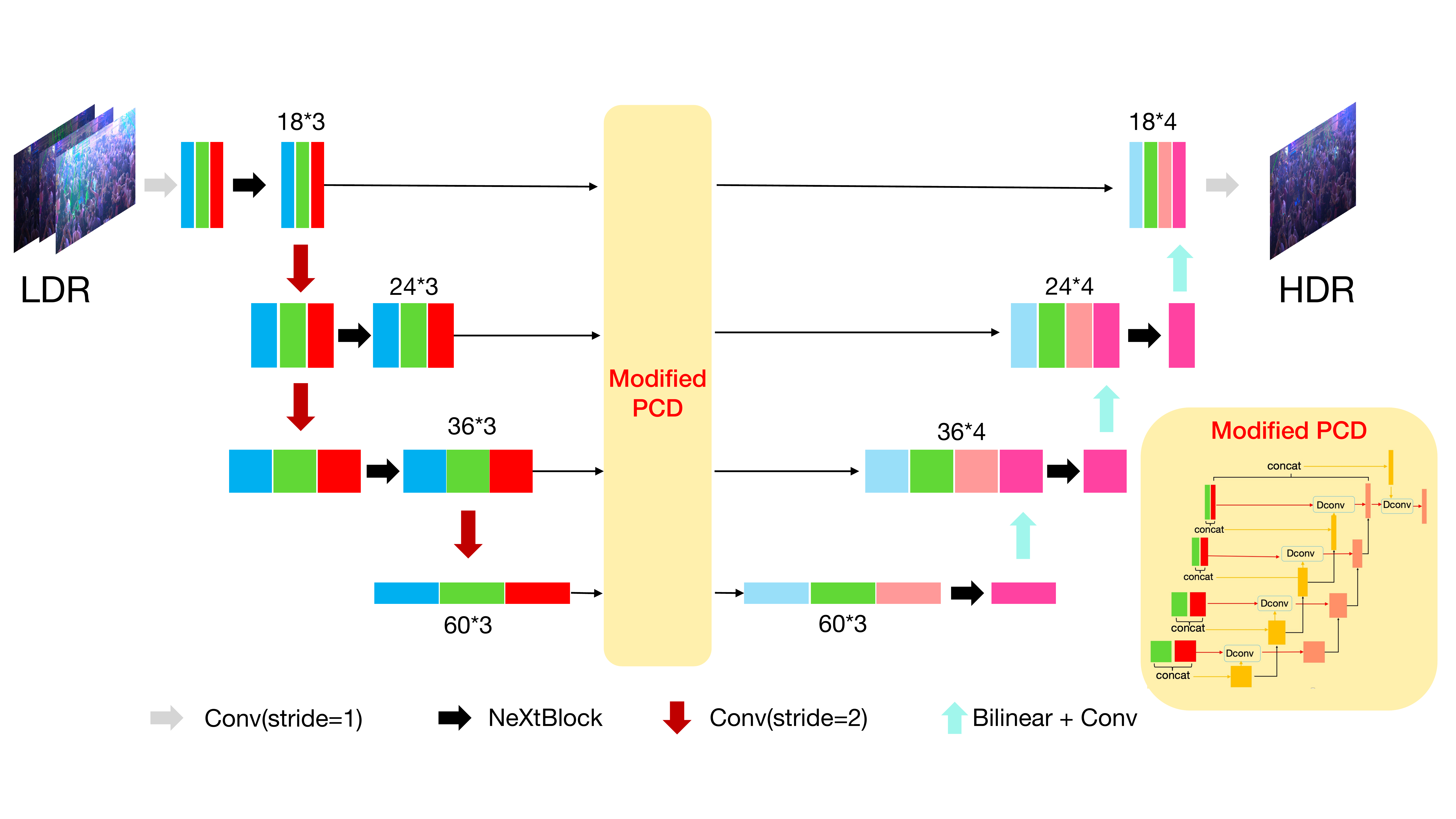}
\caption{Architecture of HDR Reconstruction with a Lightweight Multi-Level Attention Network,
proposed by the TVHDR team.}
\label{fig:TVHDR} 
\end{figure}

The team proposes an encoder-decoder structure similar to U-Net \cite{unet}. Multi-scale features are extracted from multiple LDR images and aligned to reconstruct an HDR image. The architecture of the proposed network is illustrated in Figure~\ref{fig:TVHDR}. Firstly LDR images are aligned in terms of their relative exposure value (EV), and later fed into the encoder to extract multi-scale features. To align the multi-scale features, the team proposes a modified PCD module~\cite{wang19}, aligning the features from low resolution to high resolution. The decoder utilizes all the aligned features to obtain the HDR image. To reduce the network complexity in the modified PCD module, conventional convolutions are replaced by depth-wise and point-wise convolutions, and the conventional convolutions in the encoder and decoder are replaced with ConvNeXt blocks~\cite{liu2022convnet}.

\subsection{ForLight}
\begin{figure}[h]
\centering
\includegraphics[width=1.0\columnwidth]{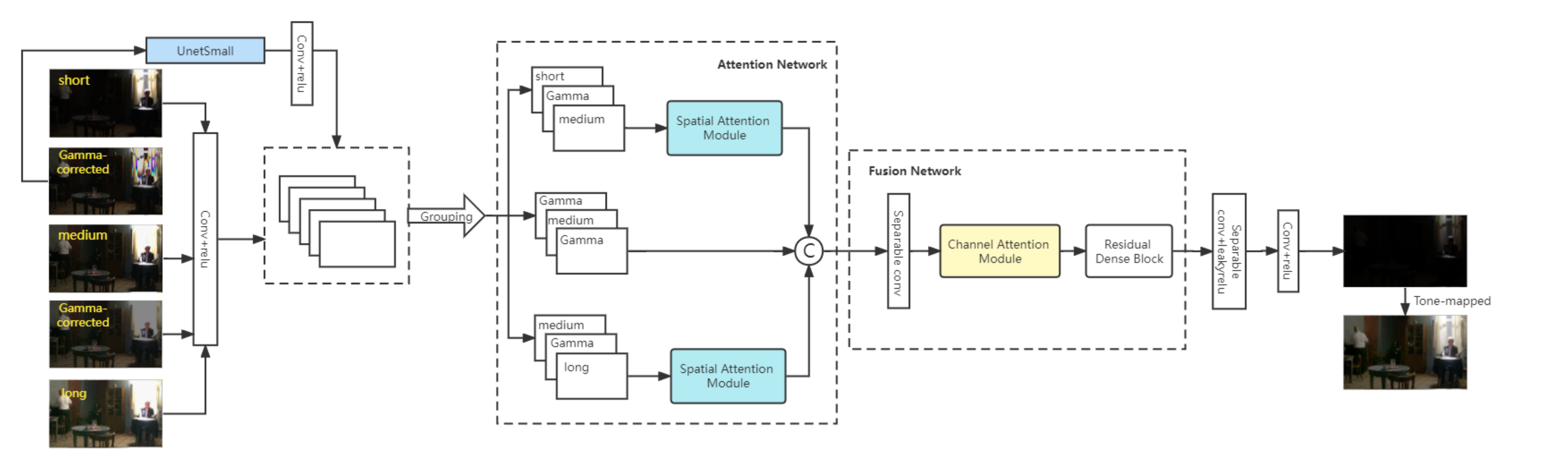}
\caption{Architecture of HDR Reconstruction with Gamma-enhanced Spatial Attention Network, proposed by the ForLight team.}
\label{fig:ForLight} 
\end{figure}

The team proposed GSANet (Gamma-enhanced Spatial Attention Network) for HDR reconstruction~\cite{fangyali2022ntire}. An overview of the two-stage framework architecture is shown in Figure~\ref{fig:ForLight}. The first stage performs feature extraction. The second stage aims for feature fusion and HDR  reconstruction. 
To remove noise and ghosting effects, the gamma transformed input is processed by a small U-Net, resulting in \emph{denoised} features. The features then undergo a spatial attention module (as proposed by~\cite{yan19}) and are finally fused
via a channel attention~\cite{woo2018cbam} and a dilated residual dense block.

\subsection{IMVIA}
\begin{figure}[h]
\centering
\includegraphics[width=1.0\columnwidth]{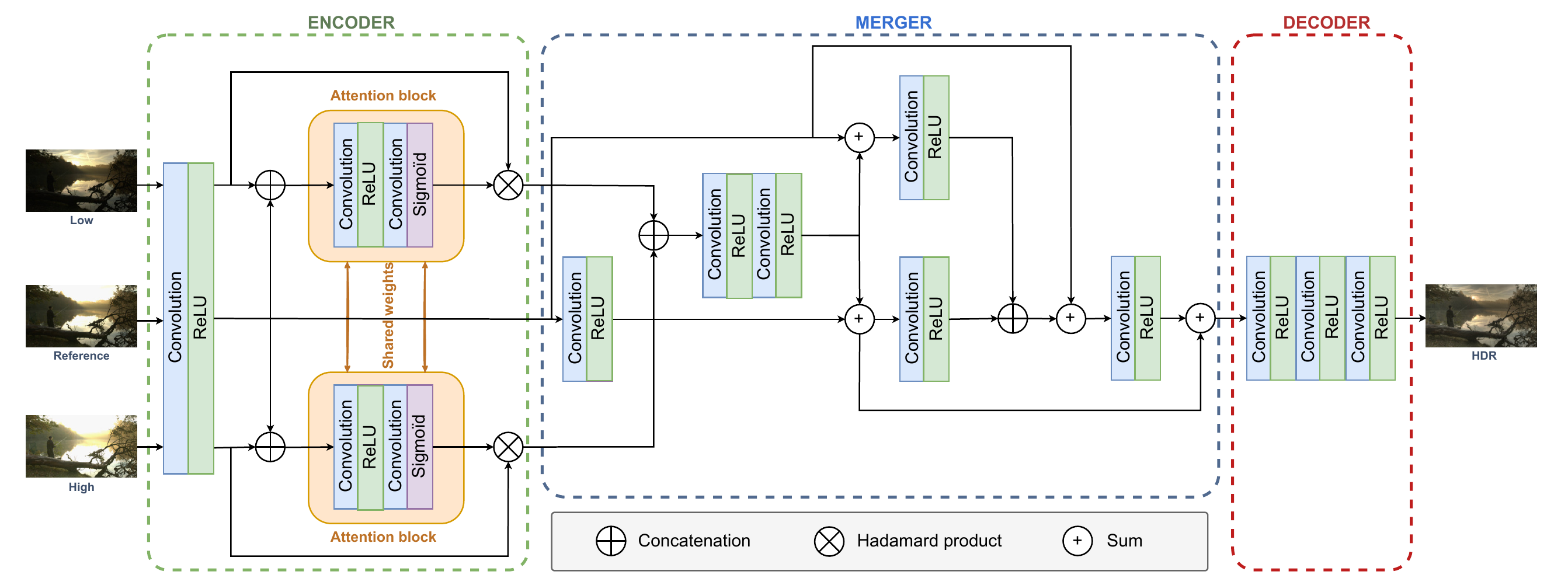}
\caption{Architecture of HDRES: Lightweight Network for Ghost-free High Dynamic Range
Imaging on Embedded System, proposed by the IMVIA team.}
\label{fig:IMVIA} 
\end{figure}

The team presents HDRES, a network with low computational cost using a novel merger architecture based on a cascading channel-spatial attention operation enabling a low number of parameters (38K). The architecture of the proposed network is illustrated in Figure~\ref{fig:IMVIA}. The encoder network is limited to only 16 feature maps to keep a low computational cost. Following previous work~\cite{yan19} attention maps are generated for each non-reference input. However, unlike~\cite{yan19} the attention blocks share the same weights to limit the number of weights in the network. The merger network firstly merges the non-reference tensors and then merges this non-reference tensor with the reference tensor. The merger architecture consists of cascading operations of sum and concatenation, offering a solution of spatial and channel attention with low computational cost. Residual connections are added to the reference tensor to facilitate the network's training. A simple decoder network, composed of 3 sequential convolution operations, produces the final HDR image from the merger's 16-channel output.

\subsection{DGU-CILAB}
\begin{figure}[h]
\centering
\includegraphics[width=1.0\columnwidth]{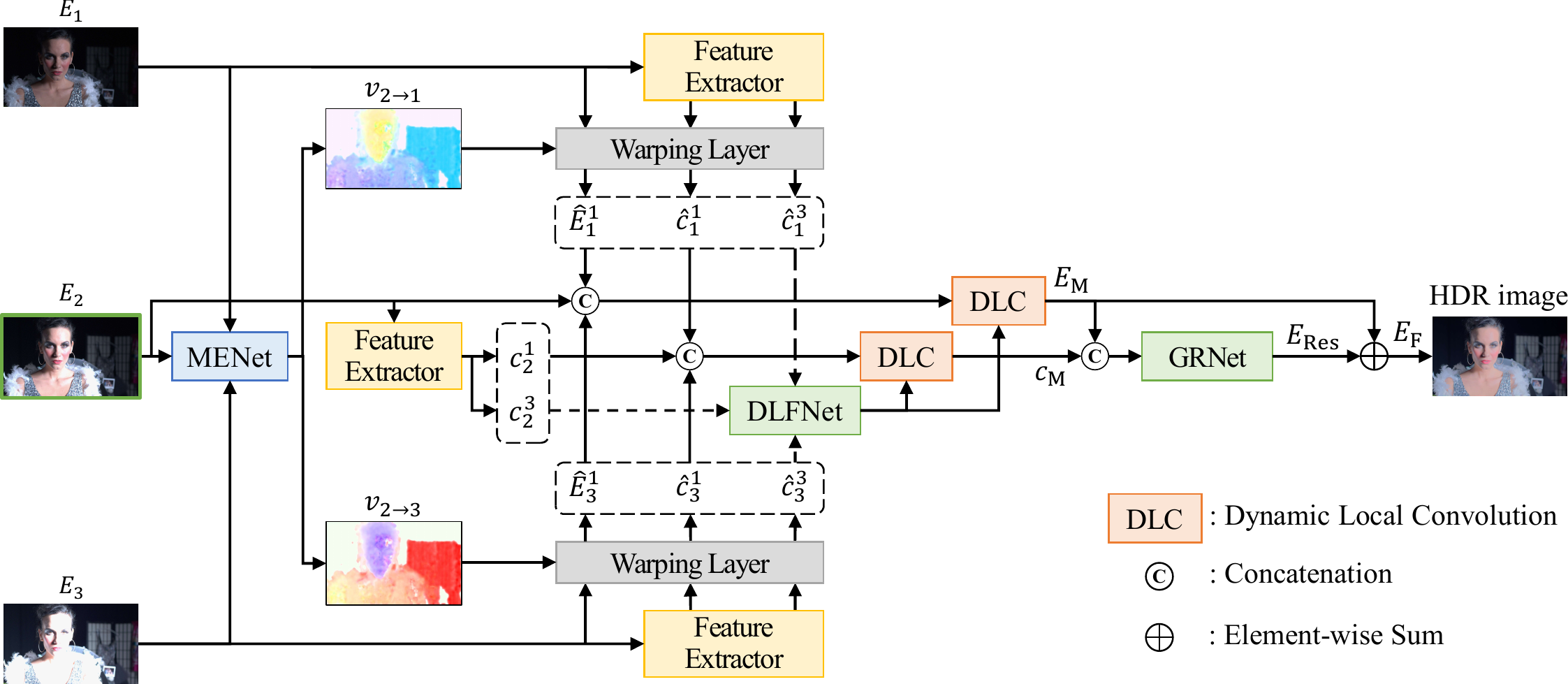}
\caption{Architecture of Bidirectional Motion Estimation with Cyclic Cost Volume for High Dynamic Range Imaging, proposed by the DGU-CILAB team.}
\label{fig:DGU-CILAB} 
\end{figure}

The team propose a learning-based HDR imaging algorithm based on bidirectional motion estimation~\cite{angiavien2022ntire}. The architecture of the proposed network is illustrated in Figure~\ref{fig:DGU-CILAB}. First, the motion estimation network (MENet) with cyclic cost volume and spatial attention maps estimates accurate optical flows between input LDR images. Then, the dynamic local fusion network (DLFNet) combines the warped and reference inputs to generate a synthesized output by exploiting local information. Finally, a global refinement network (GRNet) generates a residual image by using global information to improve the synthesis performance further.

\subsection{CVIP}
\begin{figure}[h]
\centering
\includegraphics[width=1.0\columnwidth]{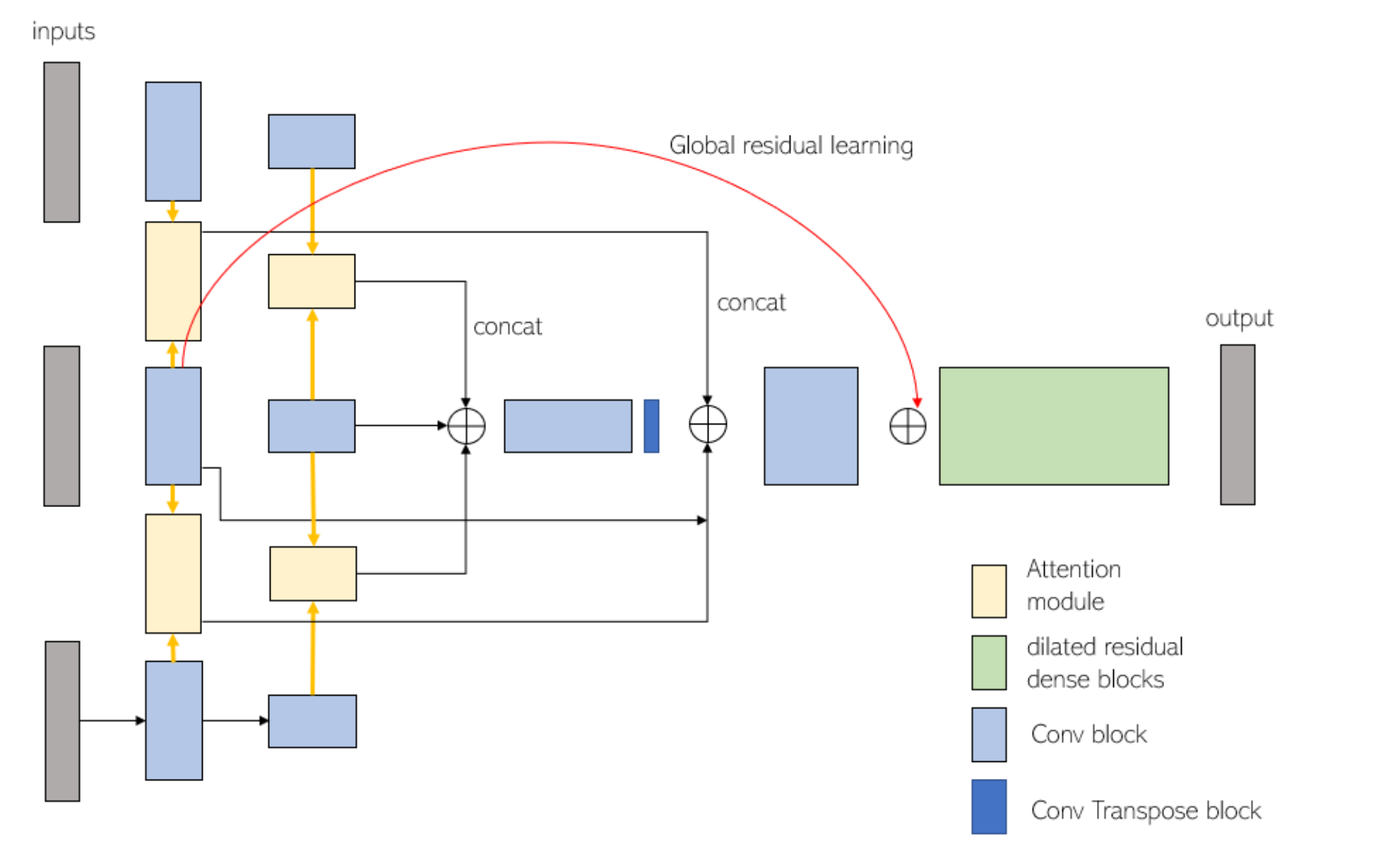}
\caption{Architecture of Multi Level Attention U-Net, proposed by the CVIP team.}
\label{fig:CVIP} 
\end{figure}

The team proposes Multi-Level Attention U-Net (MLAUNet) inspired by AHDRNet~\cite{yan19}. The architecture of the proposed network is illustrated in Figure~\ref{fig:CVIP}. The overall structure of MLAUNet primarily consists of two components: 1) an LDR encoder using a spatial attention module that identifies beneficial features from the long and short exposure frames, and 2) a decoder for final HDR reconstruction. First, the encoder successively downsamples the input frames into two levels, and the spatial attention module extracts attention feature maps at each level. Then, a decoder upsamples the feature maps to full resolution using skip-connections between each level of concatenated attention feature maps. The features from each level are merged using a convTranspose block and are passed through a series of residual dense blocks (DRDBs) to generate the final HDR image with a global residual learning strategy. 

\subsection{KCML2}
\begin{figure}[h]
\centering
\includegraphics[width=1.0\columnwidth]{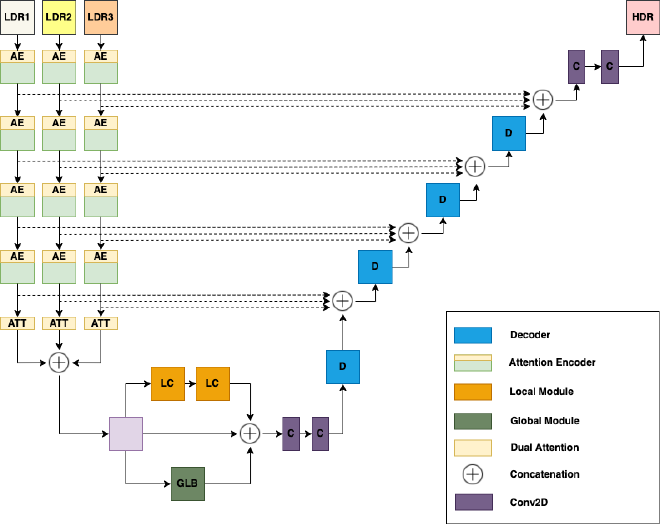}
\caption{Architecture of Attention-guided Non-Local Network (ATNLC) for High Dynamic Range Image Synthesis, proposed by the KCML2 team.}
\label{fig:KCML2} 
\end{figure}

The team presents an attention-guided non-local network (ATNLC) to deal with possible motion between LDR frames. The attention mechanism weights the contribution of each input LDR, while the non-local mechanism works for motion removal. The architecture of the proposed network is illustrated in Figure~\ref{fig:KCML2}. The proposed ATNLC comprises four modules: attention encoder, local, global, and decoder. The attention encoder, applied to each input LDR at every stage, includes a Dual-Attention (DA) module, which in turn consists of a Spatial Attention (SA) and Channel Attention (CA) modules, evaluating the importance of each pixel and channel, respectively. The local module extracts local features at different scales with varying kernel sizes, while the global module spans a large receptive field. Finally, the HDR image is obtained by employing several decoder modules; the output of each decoder is concatenated with the masked-encoded features from the attention encoder before being fed to the next decoder.

\section*{Acknowledgments}
We thank the NTIRE 2022 sponsors: Huawei, Reality Labs, Bending Spoons, MediaTek, OPPO, Oddity, Voyage81, ETH Z\"urich (Computer Vision Lab) and University of W\"urzburg (CAIDAS).

\appendix\section{Teams and Affiliations}
\label{ap:teams-and-affiliations}
\subsection*{NTIRE 2022 organizers}

\noindent \textbf{Title:} NTIRE 2022 Challenge on High Dynamic Range Imaging 
\newline
\textbf{Members:} 
Eduardo P\'erez-Pellitero$^1$\\(\href{e.perez.pellitero@huawei.com}{e.perez.pellitero@huawei.com}), Sibi Catley-Chandar$^1$, Richard Shaw$^1$, Ale\v{s} Leonardis$^1$, Radu Timofte$^2$
\newline
\textbf{Affiliations:} $^{1}$ Huawei Noah's Ark Laboratory\\
$^2$ University of W\"urzburg and ETH Z\"urich

\subsection*{ALONG}
\noindent \textbf{Title:} EfficientHDR: Residual Feature Network and Transformer Teacher for Efficient High Dynamic Range  
\newline
\textbf{Members:} Zexin Zhang (\href{zhangzexin@corp.netease.com}{zhangzexin@corp.netease.com}), Cen Liu, Yunbo Peng, Yue Lin
\newline
\textbf{Affiliations:} Netease Games AI Lab, China

\subsection*{Antins\textunderscore cv}
\noindent \textbf{Title:} Efficient Progressive High Dynamic Range Image Restoration via Attention and Alignment Network 
\newline
\textbf{Members:} Gaocheng Yu (\href{yugaocheng.ygc@antgroup.com}{yugaocheng.ygc@antgroup.com}), Jin Zhang, Zhe Ma, Hongbin Wang\newline
\textbf{Affiliations:} AntGroup

\subsection*{XPixel-UM}
\noindent \textbf{Title:} SwinHDR: Efficient Swin Transformer for High Dynamic Range Imaging 
\newline
\textbf{Members:} Xiangyu Chen$^{1,2}$ (\href{chxy95@gmail.com}{chxy95@gmail.com}), Xintao Wang$^{3}$, Haiwei Wu$^{1}$, Lin Liu$^{4}$, Chao Dong$^{2}$, Jiantao Zhou$^{1}$\newline
\textbf{Affiliations:} $^{1}$ University of Macau, $^{2}$ Shenzhen Institutes of Advanced Technology, CAS, $^{3}$ Tencent PCG, $^{4}$ University of Science and Technology of China

\subsection*{AdeTeam}
\noindent \textbf{Title:} A Lightweight Network for High Dynamic Range Imaging 
\newline
\textbf{Members:} 
Qingsen Yan $^{1}$ (\href{qingsenyan@gmail.com}{qingsenyan@gmail.com}), Song Zhang$^{2}$, Weiye Chen$^{2}$, Yuhang Liu$^{1}$, Zhen Zhang$^{1}$, Yanning Zhang$^{3}$, Javen Qinfeng Shi$^{1}$, Dong Gong$^{4}$\newline
\textbf{Affiliations:} $^{1}$ The University of Adelaide, $^{2}$ Xidian University, $^{3}$ Northwestern Polytechnical University, $^{4}$ The University of New South Wales

\subsection*{BOE-IOT-AIBD}
\noindent \textbf{Title:} Multi-Level Attention U-Net for HDR Reconstruction
\newline
\textbf{Members:} Dan Zhu (\href{zhudan@boe.com.cn}{zhudan@boe.com.cn}), Mengdi Sun, Guannan Chen\newline
\textbf{Affiliations:} BOE Technology Group Co., Ltd., China

\subsection*{CZCV}
\noindent \textbf{Title:} Multi-frame High Dynamic Range Image Reconstruction using Transformer 
\newline
\textbf{Members:} Yang Hu (\href{huyang@czur.com}{huyang@czur.com}), Haowei Li, Baozhu Zou\newline
\textbf{Affiliations:} CZUR Technology Group Co., Ltd., China

\subsection*{MegHDR}
\noindent \textbf{Title:} Multi-stage Feature Distillation Network for Efficient High Dynamic
Range Imaging 
\newline
\textbf{Members:} Zhen Liu (\href{liuzhen03@megvii.com}{liuzhen03@megvii.com}), Wenjie Lin, Ting Jiang, Chengzhi Jiang, Xinpeng Li, Mingyan Han,
Haoqiang Fan, Jian Sun and Shuaicheng Liu \newline
\textbf{Affiliations:} Megvii Technology

\subsection*{0noise}
\noindent \textbf{Title:} DRHDR: Dual Branch Residual Network for Multi-Bracket High Dynamic Range Imaging 
\newline
\textbf{Members:} Juan Mar\'in-Vega$^{1,2}$ (\href{marin@imada.sdu.dk}{marin@imada.sdu.dk}), Michael Sloth$^{2}$, Peter Schneider-Kamp$^{1}$, Richard Röttger$^{1}$\newline
\textbf{Affiliations:} $^{1}$ Department of Mathematics and Computer Science (IMADA), University of Southern Denmark, $^{2}$ Esoft Systems

\subsection*{VAlgo}
\noindent \textbf{Title:} A UNet-style Network with Spatial Attention for High Dynamic Range Imaging
\newline
\textbf{Members:} Chunyang Li (\href{lichunyang6@xiaomi.com}{lichunyang6@xiaomi.com}), Long Bao\newline
\textbf{Affiliations:} Xiaomi

\subsection*{Winterfell}
\noindent \textbf{Title:} Condition Attention-guided Reconstruction Network with Multi-exposure Fusion for High Dynamic Range Images 
\newline
\textbf{Members:} Gang He$^{1}$ (\href{ghe@xidian.edu.cn}{ghe@xidian.edu.cn}), Ziyao Xu$^{1}$, Li Xu$^{1}$, Gen Zhan$^{2}$, Ming Sun$^{3}$, Xing Wen$^{3}$, Junlin Li$^{2}$\newline
\textbf{Affiliations:} $^{1}$ Xidian University, $^{2}$ ByteDance, $^{3}$ Kuaishou Technology

\subsection*{ACALJJ32}
\noindent \textbf{Title:} Multi-scale Asymmetric Learning for High Dynamic Range Imaging
\newline
\textbf{Members:} Jinjing Li$^{1}$ (\href{1792418414@qq.com}{1792418414@qq.com}), Chenghua Li$^{2}$, Ruipeng Gang$^{3}$, Fangya Li$^{1}$, Chenming Liu$^{3}$, Shuang Feng$^{1}$\newline
\textbf{Affiliations:} $^{1}$ Communication University of China, $^{2}$ Institute of Automation, Chinese Academy of Sciences, $^{3}$ Academy of Broadcasting Science, NRTA

\subsection*{WorkFromHome}
\noindent \textbf{Title:} Deep Residual U-Net for High Dynamic Range Imaging 
\newline
\textbf{Members:} Fei Lei (\href{leifei217@gmail.com}{leifei217@gmail.com}), Rui Liu, Junxiang Ruan\newline
\textbf{Affiliations:} Tetras AI Technology 

\subsection*{TeamLiangJian}
\noindent \textbf{Title:} Wavelet-based Neural Network for Efficient High Dynamic Range Imaging 
\newline
\textbf{Members:} Tianhong Dai$^{1}$ (\href{tianhong.dai15@imperial.ac.uk}{tianhong.dai15@imperial.ac.uk}), Wei Li$^{2}$, Zhan Lu$^{2}$, Hengyan Liu$^{1}$, Peian Huang$^{3}$, Guangyu Ren$^{1}$\newline
\textbf{Affiliations:} $^{1}$ Imperial College London, $^{2}$ Tsinghua University, $^{3}$ The University of Edinburgh

\subsection*{TVHDR}
\noindent \textbf{Title:} HDR Reconstruction with a Lightweight Multi-Level Attention Network 
\newline
\textbf{Members:} Yonglin Luo$^{1}$ (\href{luoylin2007@126.com}{luoylin2007@126.com}), Chang Liu$^{2}$, Qiang Tu$^{3}$ \newline
\textbf{Affiliations:} $^{1}$ Sun Yat-sen University, $^{2}$ Shanghai Jiao Tong University, $^{3}$ Beijing Institute of Technology

\subsection*{ForLight}
\noindent \textbf{Title:} Gamma-enhanced Spatial Attention Network for Efficient High Dynamic Range Imaging
\newline
\textbf{Members:} Fangya Li$^{1}$ (\href{fly72109@163.com}{fly72109@163.com}), Ruipeng Gang$^{2}$, Chenghua Li$^{3}$, Jinjing Li$^{1}$, Sai Ma$^{2}$, Chenming Liu$^{2}$, Yizhen Cao$^{1}$\newline
\textbf{Affiliations:} $^{1}$ Communication University of China, $^{2}$ Academy of Broadcasting Sciencience, NRTA, $^{3}$ Institute of Automation, Chinese Academy of Science

\subsection*{IMVIA}
\noindent \textbf{Title:} HDRES: Lightweight Network for Ghost-free High Dynamic Range Imaging on Embedded System
\newline
\textbf{Members:} Steven Tel (\href{steven.tel@u-bourgogne.fr}{steven.tel@u-bourgogne.fr}), Barthelemy Heyrman, Dominique Ginhac\newline
\textbf{Affiliations:} ImViA laboratory, University of Burgundy

\subsection*{DGU-CILAB}
\noindent \textbf{Title:} Bidirectional Motion Estimation with Cyclic
Cost Volume for High Dynamic Range Imaging 
\newline
\textbf{Members:} Chul Lee (\href{chullee@dongguk.edu}{chullee@dongguk.edu}), Gahyeon Kim, Seonghyun Park, An Gia Vien, Truong Thanh Nhat Mai\newline
\textbf{Affiliations:} Department of Multimedia Engineering, Dongguk University

\subsection*{CVIP}
\noindent \textbf{Title:} Multi Level Attention U-Net
\newline
\textbf{Members:} Howoon Yoon (\href{hannah1258@naver.com}{hannah1258@naver.com}) 
\newline
\textbf{Affiliations:}  Gachon university

\subsection*{KCML2}
\noindent \textbf{Title:} An Attention Non Local Network for High Dynamic Range Image Synthesis
\newline
\textbf{Members:} Tu Vo (\href{tuvv@kc-ml2.com}{tuvv@kc-ml2.com}), Alexander Holston, Sheir Zaheer, Chan Y. Park 
\newline
\textbf{Affiliations:} KC Machine Learning Lab

{\small
\bibliographystyle{ieee_fullname}
\bibliography{egbib}
}

\end{document}